\documentclass[10pt,journal,compsoc]{IEEEtran}
\usepackage{cite}
\usepackage{amsmath,amssymb,amsfonts}
\usepackage{algorithm,algorithmic}
\usepackage{graphicx}
\usepackage{textcomp}
\usepackage{comment}
\usepackage{threeparttable}
\usepackage{hyperref}
\usepackage{threeparttable}
\usepackage{multirow}
\usepackage{booktabs}
\usepackage{subfig}

\def\BibTeX{{\rm B\kern-.05em{\sc i\kern-.025em b}\kern-.08em
    T\kern-.1667em\lower.7ex\hbox{E}\kern-.125emX}}
    
\begin{document}
\title{DS-MS-TCN: Otago Exercises Recognition with a Dual-Scale Multi-Stage Temporal Convolutional Network}
\author{Meng~Shang,
        Lenore~Dedeyne,
        Jolan~Dupont,
		Laura~Vercauteren,
		Nadjia~Amini,
		Laurence~Lapauw,
		Evelien~Gielen,
		Sabine~Verschueren,
		Carolina~Varon,
		Walter~De Raedt,
        and~Bart~Vanrumste
\thanks{M. Shang, C. Varon and B. Vanrumste are with KU Leuven, STADIUS, Department of Electrical Engineering, 3000 Leuven, Belgium, e-mail: meng.shang@kuleuven.be.}
\thanks{M. Shang and W. De Raedt are with Imec, Kapeldreef 75, 3001 Leuven, Belgium.}
\thanks{M. Shang and B. Vanrumste are with KU Leuven, e-Media Research lab.}
\thanks{L. Dedeyne, J. Dupont, L. Vercauteren, N. Amini, L. Lapauw, E. Gielen are with Geriatrics \& Gerontology, Department of Public Health and Primary Care, KU Leuven, Belgium.}
\thanks{J. Dupont and E. Gielen are with Department of Geriatric Medicine, UZ Leuven, Belgium.}
\thanks{S. Verschueren is with the Musculoskeletal Rehabilitation Research Group, Department of Rehabilitation Sciences, KU Leuven.}
}

\maketitle

\begin{abstract}
The Otago Exercise Program (OEP) represents a crucial rehabilitation initiative tailored for older adults, aimed at enhancing balance and strength. Despite previous efforts utilizing wearable sensors for OEP recognition, existing studies have exhibited limitations in terms of accuracy and robustness. This study addresses these limitations by employing a single waist-mounted Inertial Measurement Unit (IMU) to recognize OEP exercises among community-dwelling older adults in their daily lives. A cohort of 36 older adults participated in laboratory settings, supplemented by an additional 7 older adults recruited for at-home assessments. The study proposes a Dual-Scale Multi-Stage Temporal Convolutional Network (DS-MS-TCN) designed for two-level sequence-to-sequence classification, incorporating them in one loss function. In the first stage, the model focuses on recognizing each repetition of the exercises (micro labels). Subsequent stages extend the recognition to encompass the complete range of exercises (macro labels). The DS-MS-TCN model surpasses existing state-of-the-art deep learning models, achieving f1-scores exceeding 80\% and Intersection over Union (IoU) f1-scores surpassing 60\% for all four exercises evaluated. Notably, the model outperforms the prior study utilizing the sliding window technique, eliminating the need for post-processing stages and window size tuning. To our knowledge, we are the first to present a novel perspective on enhancing Human Activity Recognition (HAR) systems through the recognition of each repetition of activities.
\end{abstract}


\section{Introduction}
\label{sec:introduction}
\IEEEPARstart{H}uman activity recognition (HAR) has emerged as a significant research area in recent years due to its wide range of applications in various domains, including healthcare \cite{wang_survey_2019}, sports monitoring \cite{mekruksavanich2022multimodal}, and smart environments \cite{bianchi2019iot}. Wearable sensors, such as inertial measurement units (IMUs), are popular for HAR since they provide a non-invasive and unobtrusive means of capturing human motion and activity patterns. HAR systems focus on developing machine learning algorithms and systems capable of automatically identifying and classifying human activities based on sensor data.\par

The Otago Exercise Program (OEP) comprises a set of activities designed to enhance balance, strength, and walking abilities, thereby mitigating the risk of falls in older adults \cite{thomas_does_2010}. Participants are instructed to engage in these exercises two or three times a week over consecutive weeks \cite{mat_effect_2018,almarzouki_improved_2020}. To overcome the drawbacks of self-reports (diaries), researchers have proposed HAR systems using wearable IMUs. In our previous study \cite{shang2023otago}, a hierarchical system was proposed and two tasks were performed: first, the OEP was recognized from Activities of Daily Life (ADLs) with f1-scores over 0.95, and second, some OEP sub-classes were recognized with f1-scores over 0.8. However, the systems required post-processing stages to smooth the output, and the window sizes needed to be tuned.\par

Many dynamic human activities consist of repetitive movements (e.g., walking, eating, cycling). To date, much of the research up to now has annotated the daily activities and repetitive exercises as a whole \cite{wang_survey_2019, dang2020sensor}. In other words, all movements belonging to the same activities were given the same ground truth label. However, this method also annotates some irrelevant movements/gestures of the activity. For example, \textit{chair rising} as a common rehabilitation exercise, requires the repetition of sitting down and standing up. However, there are some in-between static sitting and standing gestures that are irrelevant. These irrelevant movements/gestures are also labeled as \textit{chair rising} although they are actually not distinguishable. For food intake recognition \cite{kyritsis2017food}, it has been proven possible to only annotate and recognize each cycle of a single gesture (micro movements) based on seq-to-seq classification. However, for HAR, such an annotation method has not been explored much. \par

The aim of this study is to develop a robust HAR system for OEP recognition for community-dwelling older adults. This study proposes a Dual-Scale MS-TCN (DS-MS-TCN) model with an innovative method to label and classify sensor data, called micro labeling. During annotation, each sample was given two labels: a label for each repetition of activities (micro label) and a label for the whole period of activities (macro label). Then the proposed DS-MS-TCN was applied to classify the labels. The first stage of the DS-MS-TCN learns from the raw sensor data and classifies micro labels. The following stages only learn from the output probability of the previous stage and classify macro labels. The system was applied to predict OEP for older adults using a single IMU on the waist. As an extended study of the previous work \cite{shang2023otago}, the proposed system showed outstanding results compared with traditional HAR systems. It was able to recognize some Otago exercises for older adults in their daily lives with high robustness and generalization. \par

The contributions of this study are:
\begin{itemize}
	\item This study proposes a micro-labeling method to annotate sensor data for HAR. This method annotates each repetition of the activities (micro labels), which offers less variance for the model to converge efficiently. Based on the micro labels, the whole period of activities (macro labels) could be recognized with higher accuracy compared with the same model without micro labels.
	\item This study applies a DS-MS-TCN model. The first stage classifies the micro activities while the following stages learn from the output of the previous stage and classify the macro activities. The models outperformed the other state-of-the-art deep learning models with higher f1-scores.
	\item This study proves the possibility of using a single IMU to recognize OEP in daily life. By the proposed seq-to-seq system, four OEP exercises were classified with high f1-scores and low over-segmentation error. The system could be generalized to the home settings. Compared with the previous study \cite{shang2023otago}, the system did not require sliding windows or any post-processing stages.
\end{itemize}

The paper is structured as follows: Section~\ref{sec:relatedworks} reviews the state of the art of HAR systems. Section~\ref{sec:method} outlines the datasets and describes the implementation and validation of the proposed system. Section~\ref{sec:results} presents the experimental findings, and Section~\ref{sec:discussion} delves into the discussion of these results. Lastly, Section~\ref{sec:conclusion} concludes the paper and suggests avenues for future research.

\section{Related works}
\label{sec:relatedworks}
The sliding window technique is commonly used. It segments the input sensors data into windows with equal size \cite{9153742, li_human_2022, ellis_multi-sensor_2014}. Each window with multiple samples is then classified as a certain class (window-wise classification). The drawback is that each window is independent of the others during training and the model thus has limited knowledge of the temporal dependencies. Also, the determination of sliding window sizes requires domain knowledge and experiments for hyperparameter tuning, as in our previous study for OEP recognition \cite{shang2023otago}. On the other hand, in the field of action recognition by videos, it has been proven efficient to train the models with a time series of frames \cite{farha_ms-tcn_2019}. Inspired by this, the sensor data could also be processed with sequence-to-sequence (seq-to-seq) systems \cite{wang2022eat, shang2023multi}, which makes it possible that each sample serves as a training example (sample-wise classification). However, for rehabilitation programs, such systems still lack validation.\par

Traditional machine learning approaches have long been the cornerstone of many data analysis tasks, including HAR \cite{wang_survey_2019}. These methods typically involve the extraction of handcrafted features from raw sensor data, followed by the application of machine learning algorithms to classify activities. Feature engineering requires domain expertise and careful selection of relevant features, which can be time-consuming and may limit the model's ability to capture intricate patterns in complex datasets. In contrast, deep learning has emerged as a groundbreaking paradigm in artificial intelligence, revolutionizing various fields by automatically learning hierarchical representations directly from raw data. By leveraging architectures such as recurrent neural networks (RNNs) \cite{zhao_deep_2018} and convolutional neural networks (CNNs) \cite{8684824, lee_human_2017, wagner_activity_2017, tao2018worker}, deep learning models can effectively learn hierarchical representations, automatically extract discriminative features, and model complex temporal dependencies in activity sequences. Recently, researchers have applied the combination of CNN and Long short-term memory (LSTM) cells (CNN-LSTM) \cite{mutegeki_cnn-lstm_2020, mekruksavanich_smartwatch-based_2020}, which outperformed single CNN or LSTM network on many datasets. Such an architecture could represent both local spatial features and long-time temporal features. Besides, transformers are becoming popular for HAR systems and have demonstrated efficiency for seq-to-seq recognition \cite{dirgova2022wearable}. However, transformers did not outperform CNN-LSTM according to the previous study for IMU-based activity recognition \cite{trujillo2023accuracy}. \par

Recently, multi-stage temporal convolutional networks (MS-TCNs) were applied for action recognition by videos \cite{farha_ms-tcn_2019}. Then, there were several studies applying MS-TCNs for sensor data on food intake prediction \cite{wang2022eat} and physical activity recognition \cite{shang2023multi}. All of these studies reported the advantages of MS-TCNs over the state-of-the-art CNN-LSTM. MS-TCNs include dilated filters to extract features over a longer time without increasing computational cost. Besides, by stacking multiple stages of the same architecture, the sample-wise classification could be refined to reduce over-segmentation error. The issue of over-segmentation errors in HAR systems has not gained widespread attention. \par

\section{Materials and Methodology}
\label{sec:method}

\subsection{Data Collection}
This study received approval from the Ethics Committee Research UZ/KU Leuven (S59660 and S60763). Written informed consent was obtained from all participants prior to study participation.\par

In the experiments, community-dwelling older adults (aged 65 and older) performed modified OEP \cite{dedeyne_exploring_2021} while equipped with a McRoberts MoveMonitor+ (McRoberts B.V., Netherlands) featuring a 9-axis Inertial Measurement Unit (IMU). For user-friendliness, participants were instructed to wear the device loosely and comfortably on the waist. The original program includes multiple sub-classes that need to be performed sequentially. The detailed information of the sub-classes is discussed in \cite{shang2023otago}. Subjects also performed other ADLs. Two datasets were collected in the study in different scenarios:

\subsubsection{Lab}
The dataset was recorded in a laboratory setting, where participants followed the OEP and ADLs under the guidance of researchers certified as OEP leaders. Data for both OEP and ADLs were gathered either on two distinct days or consecutively within a single day. The ADLs encompassed walking, climbing stairs, sitting, standing, and indoor cycling.

\subsubsection{Home}
The dataset was gathered in a home environment through recorded videos. Participants independently wore the device with the camera turned on and followed a booklet containing instructions for performing the OEP. Additionally, both before and/or after the OEP sessions, subjects randomly performed ADLs out of the camera's view. Consequently, while the ADLs were not directly observed, they were still recorded as the subjects wore the device.

The recruited older adults were (pre-)sarcopenic or non-sarcopenic (defined by EWGSOP1 \cite{cruz2010sarcopenia}). The detailed information is shown in Table~\ref{tab:subinfo}. The recruited subjects of the two datasets were different. During intervals between exercises, participants were neither instructed nor monitored. They had the flexibility to practice the exercises, take a seated rest, or walk out of the camera view.

\begin{table}[!t]
\caption{Information of the subjects}
\centering
\label{tab:subinfo}
\begin{tabular}{|l|l|l|ll|}
\hline
     & number & age                                                       & \multicolumn{2}{l|}{gender \&   sarcopenia}                                                                                      \\ \hline
Lab  & 36     & \begin{tabular}[c]{@{}l@{}}79.33$\pm$\\ 5.73\end{tabular} & \multicolumn{2}{l|}{\begin{tabular}[c]{@{}l@{}}17 females (7 (pre-)sarcopenia),\\ 19   males (11 (pre-)sarcopenia)\end{tabular}} \\ \hline
Home & 7      & \begin{tabular}[c]{@{}l@{}}69.43$\pm$\\ 2.92\end{tabular} & \multicolumn{2}{l|}{\begin{tabular}[c]{@{}l@{}}4 females (2 (pre-)sarcopenia),\\ 3 males (2 (pre-)sarcopenia)\end{tabular}}      \\ \hline
\end{tabular}
\end{table}

\subsection{seq-to-seq HAR}

Traditionally, HAR systems apply the sliding window technique that takes a window of consecutive samples as input for classification. A sliding window is used to segment the continuous data stream into fixed-size windows, which are then processed as a unit. This approach then aggregates multiple samples together to extract features either manually or using deep learning models. The determination of window size is challenging because it is related to the activities and subjects. \par

With the development of deep learning models, HAR systems could apply seq-to-seq operations where the sliding window technique is not necessary. This technique is sample-wise because each sample from the sensors generates an individual output.

\subsection{micro labels}

For most human activities, each activity consists of a series of sub-movements (i.e. micro activities). Most existing studies for HAR systems annotate the full range of each activity (macro labels), i.e. from the start of the first movement to the end of the last movement \cite{shang2023otago, wang_survey_2019}. However, there are still intervals between each two micro activities. For many older adults and patients, the activities were performed at a low speed, which means that there is a break after each movement. This introduces irrelevant information for the models to learn.\par

\begin{figure}[!t]
\centering
\includegraphics[width=\columnwidth]{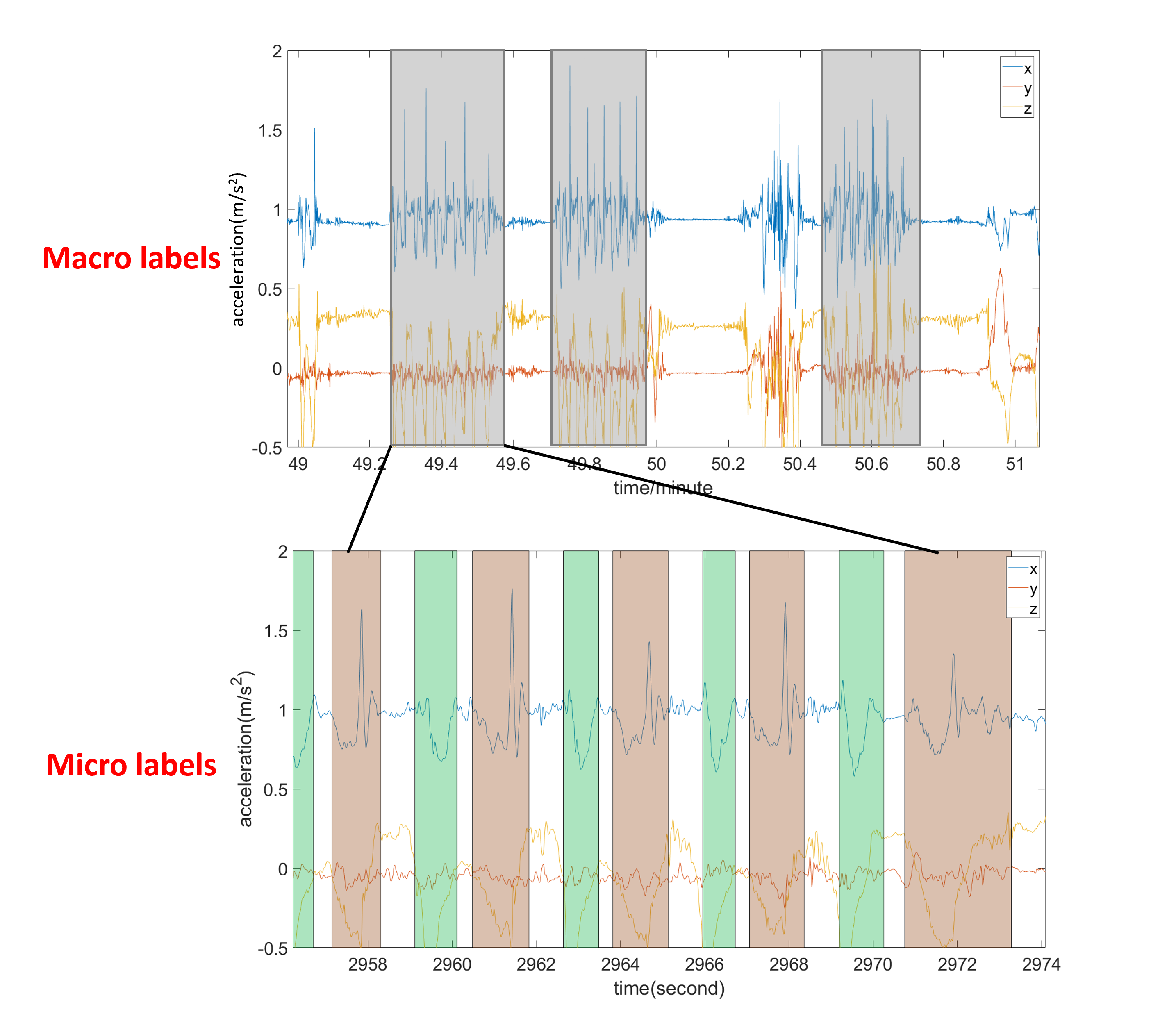}
\caption{The accelerometer signals and annotation of macro labels and micro labels for \textit{chair rising}, where each segment of macro activity consisted of five repetitions of \textit{sit-to-stand} (green blocks) and \textit{stand-to-sit} (red blocks)}
\label{fig:macrolabels}
\end{figure}

In this study, a method called micro-labeling was applied for more accurate information. Fig.~\ref{fig:macrolabels} shows an example of the micro-labeling method for \textit{chair rising}, which is an Otago Exercise. There are two types of micro activities in this activity: \textit{sit-to-stand} and \textit{stand-to-sit}. In this rehabilitation program, the subjects have to repeat these movements five times. The micro-labeling method annotates each \textit{sit-to-stand} and \textit{stand-to-sit} individually, without annotating the intervals between each repetition. In other words, a segment of macro labels consists of a set of micro labels. \par

\begin{figure*}[!t]
\centering
\subfloat[ankle plantarflexors]{\includegraphics[width=0.9\columnwidth]{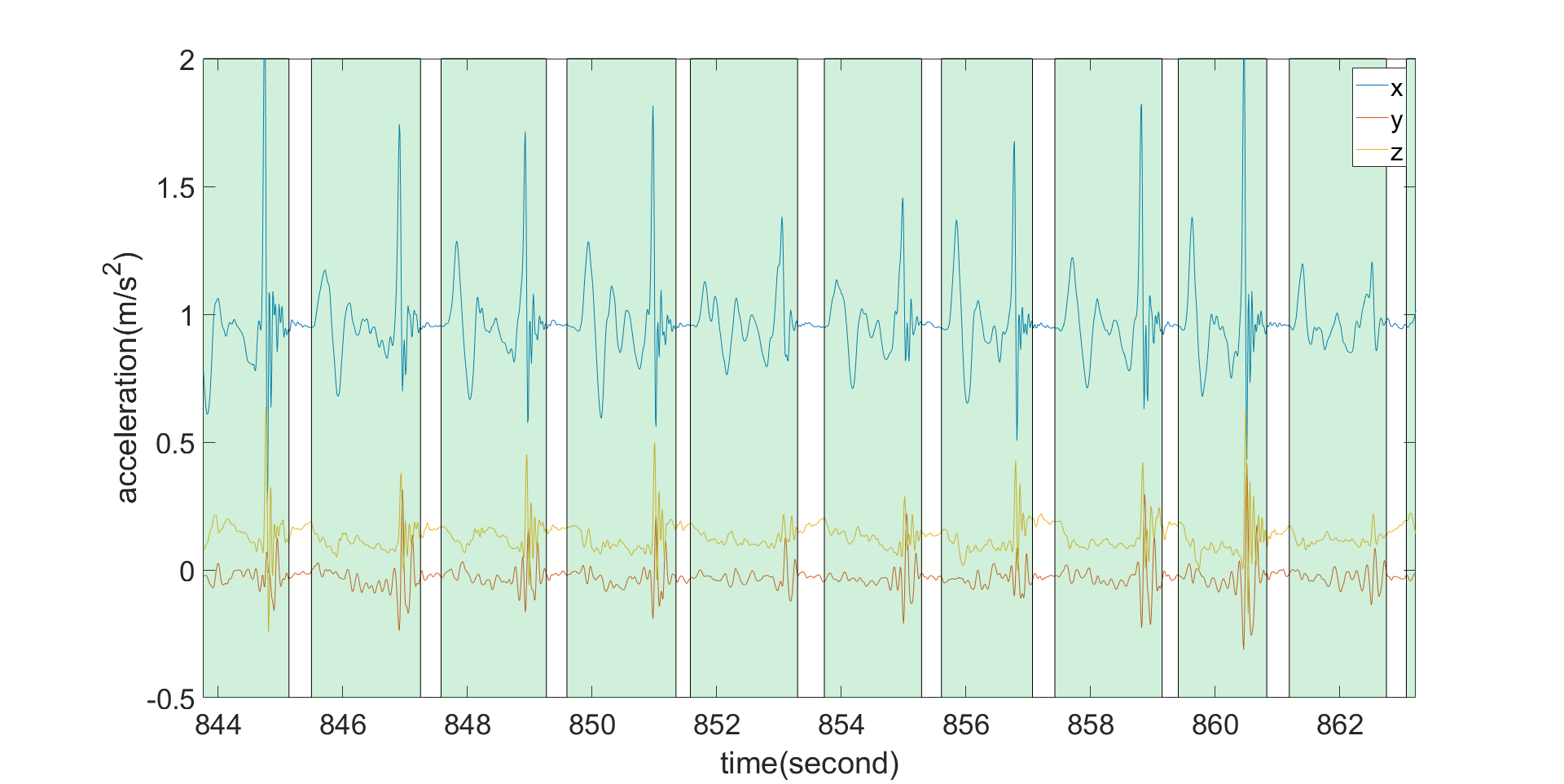}%
}
\subfloat[abdominal muscles]{\includegraphics[width=0.9\columnwidth]{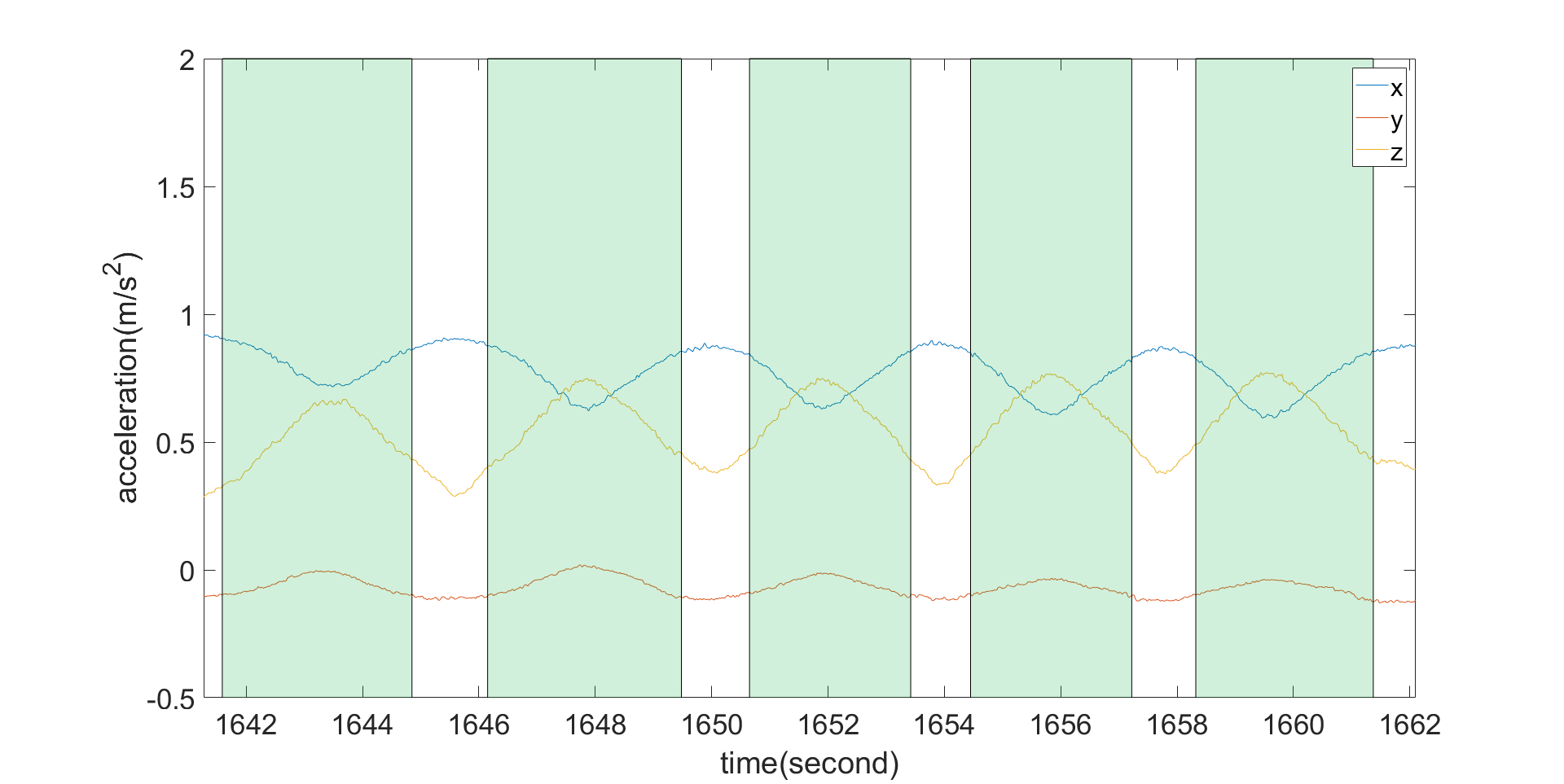}%
}
\hfill
\subfloat[knee bends]{\includegraphics[width=0.9\columnwidth]{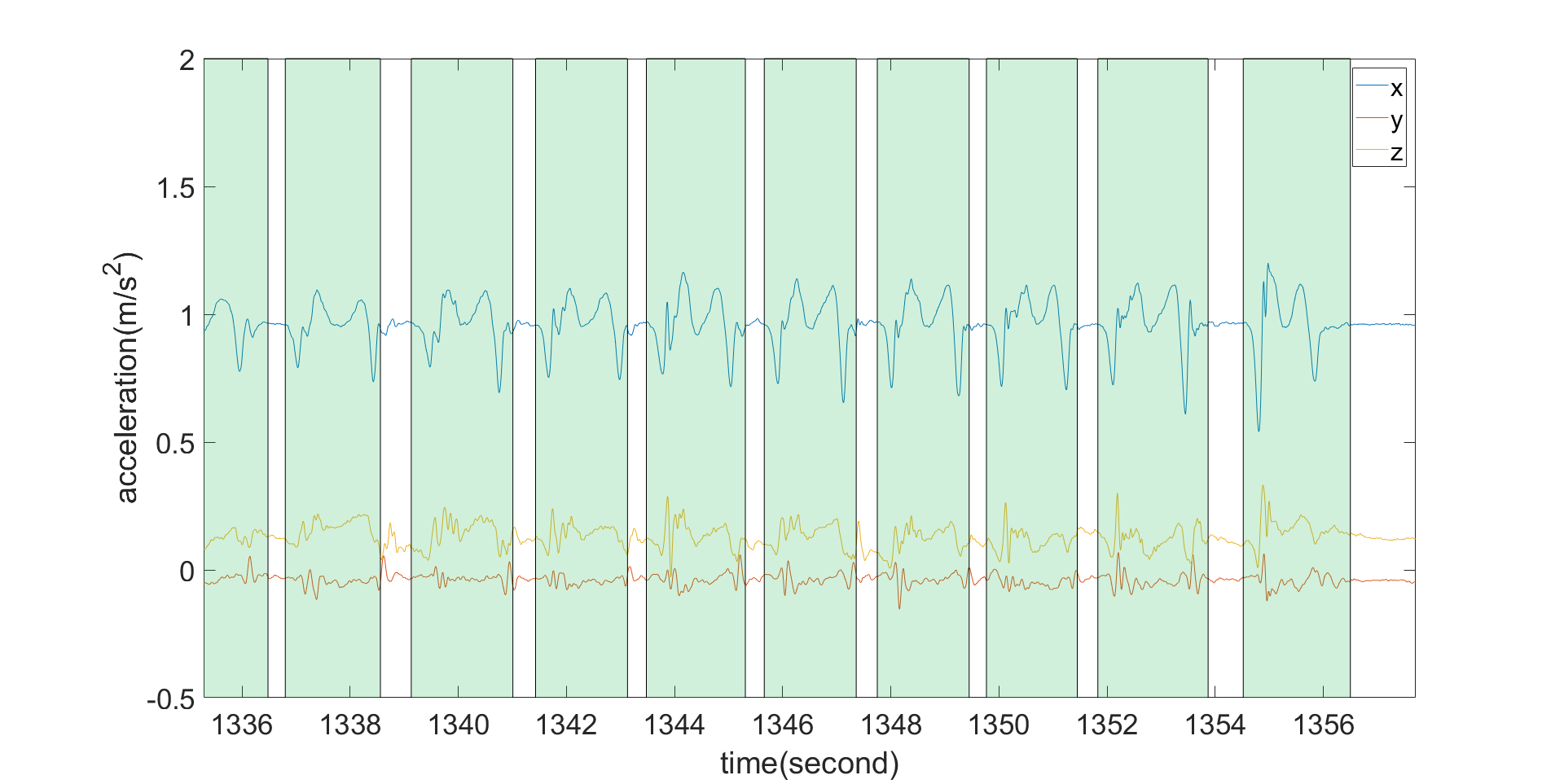}%
}
\subfloat[chair rising (green blocks = \textit{sit-to-stand}, red blocks = \textit{stand-to-sit})]{\includegraphics[ width=0.9\columnwidth]{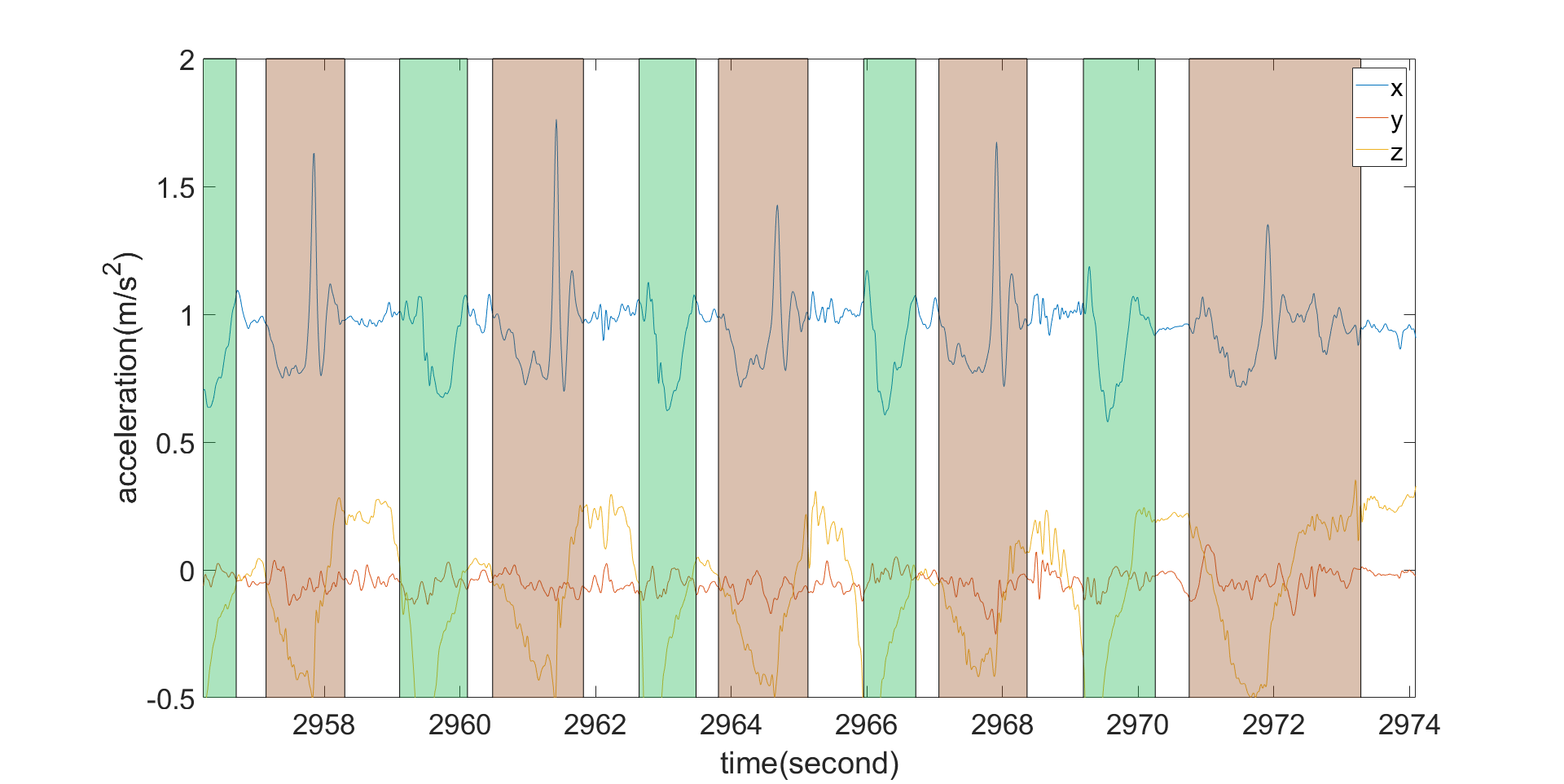}%
}
\caption{Examples of each repetition of micro activities and corresponding acceleration. The shadowed area illustrates the positive micro labels and the unshadowed area illustrates \textit{other} labels}
\label{fig:microlabels}
\end{figure*}

Similarly, each movement of \textit{ankle plantarflexors}, \textit{knee bends}, and \textit{abdominal muscles} was annotated as micro labels, as shown in Fig.~\ref{fig:microlabels}. Specifically, each segment of the micro labels was annotated according to the following observations: 

\begin{itemize}
	\item \textit{ankle plantarflexors}: start when the heels leave the ground and end when the heels fall back to the ground.
	\item \textit{knee bends}: start when the knees bend and end when the knees are straight again.
	\item \textit{abdominal muscles}: start when the trunk leans back and end when the trunk goes back to vertical.
	\item \textit{chair rising}:
		\begin{itemize}
			\item \textit{sit-to-stand}: start when the hip leaves the chair and end when standing straight.
			\item \textit{stand-to-sit}: start when the knees bend and end when taking the seat.
		\end{itemize}
\end{itemize}

As a result, there were five classes of macro labels: \textit{ankle plantarflexors}, \textit{knee bend}, \textit{abdominal muscles} \textit{chair rising}, and \textit{others}. For micro labels, correspondingly, there were six classes:  \textit{micro ankle plantarflexors}, \textit{micro knee bend}, \textit{micro abdominal muscles} \textit{sit-to-stand}, \textit{stand-to-sit}, and \textit{others}. The \textit{others} type included all other labels: ADLs, other OEP exercises, unknown activities between OEP exercises, etc. For micro labels, the intervals between repetitions were also labeled as \textit{others}. Table~\ref{tab:duration} shows the number of annotated micro and macro labeled segments duration of each class. \par

\begin{table}[!t]
\caption{The number of micro/macro labeled segments and the duration of the macro activities}
\label{tab:duration}
\centering
\begin{tabular}{llll}
\hline
                       & micro labels          & macro labels          & duration (min)          \\ \hline
ankle plan             & 970                   & 75                    & 53.09                   \\ \hline
abdominal              & 202                   & 35                    & 16.94                   \\ \hline
knee bends             & 748                   & 75                    & 33.92                   \\ \hline
chair rising           & 369*                  & 72                    & 24.11                   \\ \hline
other                  & /                     & /                     & 2483.90                 \\ \hline
\multicolumn{4}{l}{*This values is the same for \textit{sit-to-stand} and \textit{stand-to-sit}}
\end{tabular}
\end{table}

\subsection{Architecture overview: DS-MS-TCN}

\begin{figure}[!t]
\centering
\includegraphics[width=\columnwidth]{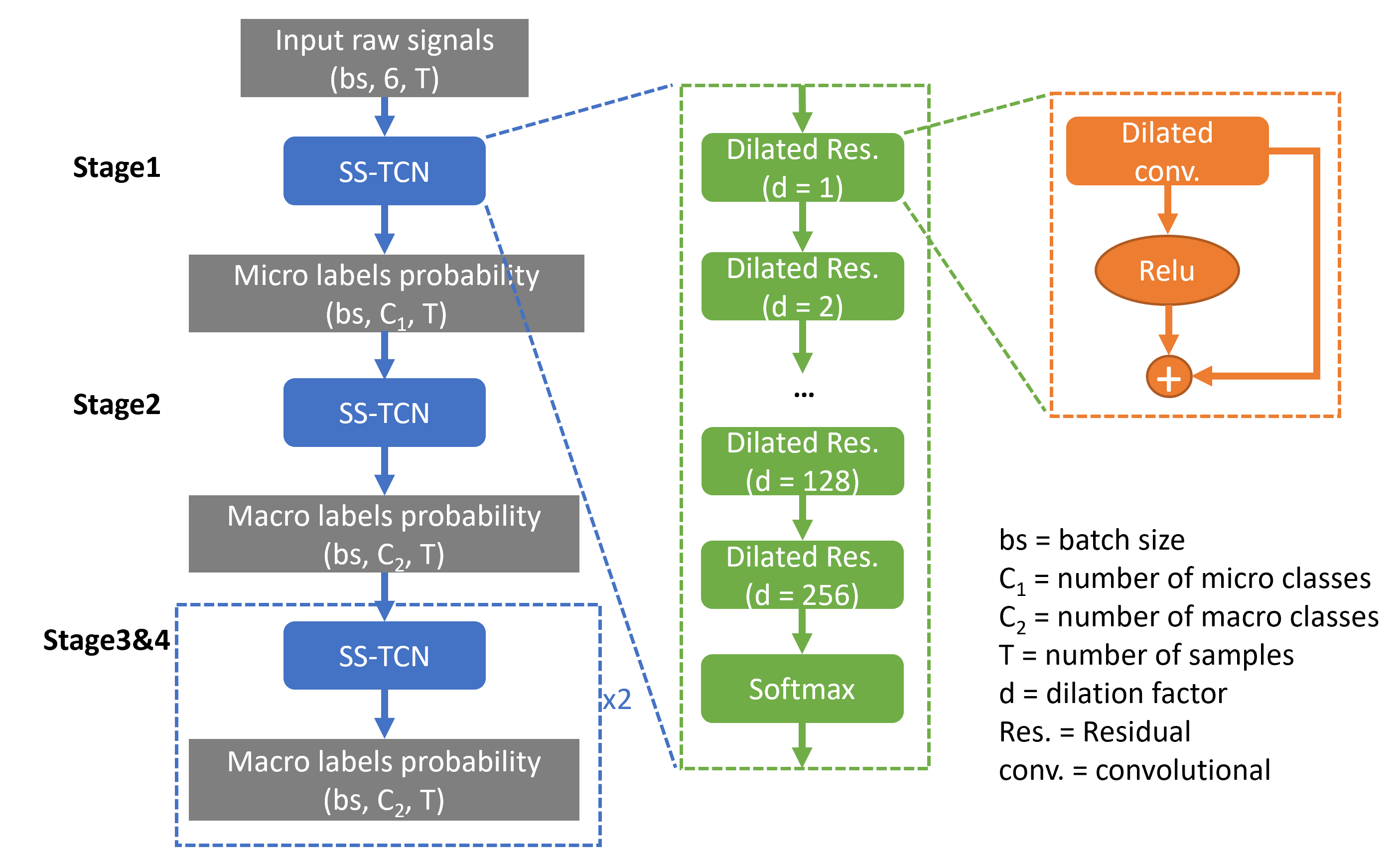}
\caption{The overview of the proposed DS-MS-TCN. Four stages of SS-TCN were applied to comply to classify micro labels and macro labels. Each SS-TCN composed of multiple residual dilated layers.}
\label{fig:overview}
\end{figure}

In this study, a Dual-scale Multi-Stage Temporal Convolutional Network (DS-MS-TCN) was proposed based on both micro labels and macro labels. As shown in Fig.~\ref{fig:overview}, the proposed deep learning model consisted of three parts: micro labels classification, macro labels classification, and macro labels refinement. The micro labels were first classified by the first stage of the model. Then the second stage generated macro labels from the micro labels. Finally, the third stage and the fourth stage refined the macro labels to reduce over-segmentation errors. The details of each stage are explained in the following sections. \par

Each stage of the model employed a Single-stage Temporal Convolutional Network (SS-TCN) architecture, comprising a stack of multiple residual dilated convolution layers. This design facilitated the model's ability to learn from broader time intervals without incurring additional computational costs. Following each dilated convolutional layer, a RELU operation is applied. To mitigate issues related to vanishing or exploding gradients, a residual connection is employed at the output, as depicted in the right part of Fig.~\ref{fig:overview}. The input signals only included the accelerometer and gyroscope, as in our previous study \cite{shang2023otago}. Therefore, there were six channels in the input. \par

To expand the receptive field, a series of individual residual dilated convolution layers were stacked. This stacking process initiated with the first layer employing a dilation factor of 1 and then doubled the dilation factors for each subsequent layer, progressively using 2, 4, 8, 16, and so on. To simplify computations, each dilated layer was equipped with a filter of size $3\times D$, where $D$ represents the number of filters. By elevating the dilation factors, the receptive field can be calculated as follows:

\begin{equation}
\text{receptive field} = 2^{L+1}-1.
\label{eq:RF}
\end{equation}

Finally, a 1$\times$1 convolutional layer with softmax function was applied to generate the probability for each sample. This study applied nine layers for each stage as proposed in \cite{farha_ms-tcn_2019}. The receptive filed was hence 1023 samples (10.23 seconds). \par

\subsection{Stage 1: Micro labels classification}

Using an SS-TCN, the first stage recognized micro activities from the input IMU signals, as follows:
\begin{equation}
\hat{Y}_{micro} = Z_1\{X\},
\label{eq:1st}
\end{equation}

Where $X$ is the input signals, and $\hat{Y}_{micro}$ is the micro labels predicted by the first stage $Z_1$. \par

The classification stage of micro activities was not expected to generate clinically applicable results, since these micro activities happened frequently in daily life. For example, the first stage cannot recognize whether a repetition of \textit{chair rising} happened in OEP or in ADLs.

\subsection{Stage 2: Macro labels classification}

The first stage produced the classification probabilities for micro labels. Subsequently, another stage of SS-TCN was employed to perform the classification of macro labels. The second stage had the same structure as the first stage, consisting of multiple residual dilated convolution layers. The micro labels probability served as the input of the second stage to generate macro labels probability as follows:

\begin{equation}
\hat{Y}_{macro}^{[2]} = Z_2\{\hat{Y}_{micro}\},
\label{eq:2nd}
\end{equation}

Where $\hat{Y}_{macro}^{[2]}$ is the predicted labels by the second stage, noted by $Z_2$.

This stage could generate the clinically applicable macro activities recognition. By considering the temporal relationship of the micro activities, this stage can recognize which micro activities were within OEP and which were not. For example, if a single \textit{chair rising} happened in daily life, the second stage would classify this event as ADLs. On the other hand, if five consecutive repetitions of \textit{chair rising} were recognized in the first stage, the second stage would classify this event as \textit{chair rising}.

\subsection{Stage 3-4: Macro labels refinement}

The macro labels were expected to perform low over-segmentation errors (explained in Section\ref{subsec:evaluation}). Therefore, another two stages of SS-TCN were applied to further refine the sample-wise macro labels as a time series. For both stages, the input and the output were both the macro labels, as follows:

\begin{equation}
\hat{Y}_{macro}^{[s]} = Z_s\{\hat{Y}_{macro}^{[s-1]}\},\;for \; s=3, 4,
\label{eq:3and4}
\end{equation}

Where $\hat{Y}_{macro}^{[s]}$ denotes the refined macro labels from stage $Z_s$. The final stage generated the classified and refined macro labels.\par

\subsection{loss function}

There are two parts of the loss function: classification loss $L_{CE}$ and refinement loss $L_{TMSE}$.\par

The first stage of DS-MS-TCN performed a classification task regarding the micro activities using cross entropy (CE) loss:

\begin{equation}
L_{CE}^{[1]}=\frac{1}{TC_1}\sum_{t,c_1}-y_{\text{micro},t,c_1}\log{\hat{y}_{\text{micro},t,c_1}},
\label{LCL1}
\end{equation}
where $y_{\text{micro},t,c_1}$ and $\hat{y}_{\text{micro},t,c_1}$ denote the true and predicted micro labels at timestamp t for micro class $c_1$ in the first stage. $C_1$ is the number of micro classes, and T is the number of samples in the time series. \par

For the following stages, the classification was performed based on the macro activities
\begin{equation}
L_{CE}^{[s]}=\frac{1}{TC_2}\sum_{t,c_2}-y_{\text{macro},t,c_2}\log{\hat{y}_{\text{macro},t,c_2}^{[s]}},\; for \;s>1,
\label{LCLs}
\end{equation}

where $y_{\text{macro},t,c_2}$ denotes the true macro labels and $\hat{y}_{\text{macro},t,c_2}^{[s]}$ denotes the predicted macro labels from stage $s$ for macro class $c_2$. $C_2$ is the number of macro classes.\par

To improve the classification performance, an additional Truncated Mean Squared Error (TMSE) function was incorporated for refinement, inspired by Farha et al. \cite{farha_ms-tcn_2019}. For each stage $s$ (except the first stage), the function $L_{TMSE}^{[s]}$ is defined as

\begin{equation}
L_{TMSE}^{[s]}=\frac{1}{TC_2}\sum_{t,c_2}(\Tilde{\Delta}_{t,c_2}^{[s]})^2, \; for \;s>1,
\label{LREs}
\end{equation}

where $\Tilde{\Delta}_{t,c_2}^{[s]}$ is the truncated difference between the loss values of two adjacent samples

\begin{equation}
\Tilde{\Delta}_{t,c_2}^{[s]}=
    \begin{cases}
      \Delta_{t,c_2}^{[s]}, & \text{if $\Delta_{t,c_2}^{[s]}\leq\tau$}  \\
      \tau, & \text{otherwise}
    \end{cases},     
\label{deltat}
\end{equation}

\begin{equation}
\Delta_{t,c_2}^{[s]} = |log{\hat{y}_{\text{macro},t,c_2}^{[s]}}-log{\hat{y}_{\text{macro},t-1,c_2}^{[s]}}|.
\label{delta}
\end{equation}

$L_{TMSE}$ was applied to reduce the segmentation error. It was thus not necessary for the first stage applied for classifying the micro labels. The reason was that the micro labels were short, and refining these labels would result in some lost information for the following stages.\par

The final loss function is computed as
\begin{equation}
L = \eta L_{CE}^{[1]} + L_{CE}^{[s]}+\lambda L_{TMSE}^{[s]}, \; for \;s>1.
\label{Loss}
\end{equation}

The values of $\lambda$ and $\tau$ have been explored in the previous study \cite{farha_ms-tcn_2019}. Therefore, this study kept the optimal values as 0.15 and 4, respectively. On the other hand, since $\eta$ was originally proposed in this study, the values were explored in the following experiments.

\subsection{Baseline methods}

\subsubsection{Hierarchical system based on sliding window}
In our previous study \cite{shang2023otago}, a hierarchical system was proposed, including two-step tasks based on sliding window techniques. In this study, we made a trade-off in selecting the window size. A large window size (10 minutes) was applied for classifying general Otago exercises and Daily activities (i.e. binary classification). Then, a small window size (6 seconds) was applied to classify the sub-classes. Besides, since the windows were classified individually, additional post-processing was needed to smooth the output labels as a time series. This system based on the sliding window technique was involved as a baseline method to compare with the proposed method. To make it comparable with the seq-to-seq methods, the classified windows were reconstructed as time series. \par

\subsubsection{CNN}

A typical CNN model proposed by \cite{tao2018worker}, where three convolutional layers with dropout and pooling layers were applied to extract features. After flattening these features, they were classified by a fully connected layer. Finally, an output layer with a softmax function was applied to generate the final prediction. This model was implemented as a seq-to-seq model.

\subsubsection{Transformer}

The model was proposed by \cite{dirgova2022wearable}. After the initial steps of normalization and position embedding, the architecture employed three encoder blocks. Each of these blocks included a multi-head attention layer and a fully-connected layer, interspersed with dropout layers for regularization. Following these blocks, classification outcomes were produced using an additional fully-connected layer. This model was implemented as a seq-to-seq model.

\subsubsection{CNN-LSTM}

The architecture and hyperparameters of the CNN-LSTM model used in this study were proposed by \cite{mekruksavanich_smartwatch-based_2020}. Therefore, the strides of the convolutional kernels were all set to one. And the LSTM layer returned the sequence rather than only the last output. Then dense layer was applied to generate the probabilities of each sample. This model was implemented as a seq-to-seq model.

\subsubsection{MS-TCN (without micro labels)}

To explore the influence of the micro labels, as a benchmark, a 4-stage MS-TCN model was applied without micro labels. Therefore, the first stage of this MS-TCN model also complied with macro activity classification and refinement. For this model, $\eta$ was set to 1 in the loss function.

\subsection{Training details}

Since the signals of Otago exercises and Daily activities were long time series, the models could not converge efficiently. Therefore, the signals were cut into slices with an equal length of 40 seconds, with 50\% overlap. The length should be larger than the receptive field and the overlap was applied to mitigate information loss at the edges of non-overlapping windows. Then batch learning was applied for training with a batch size of 32.\par

The deep learning models were developed using Pytorch and trained on an NVIDIA P100 SXM2 GPU. The Adam optimization algorithm was utilized to minimize the loss function. \par

To make the deep learning models (CNN, Transformer, CNN-LSTM) comparable with the proposed model, they also complied with the seq-to-seq classification task. The CNN model and CNN-LSTM model were implemented by padding the convolutional layers to the same length as the input.\par

\subsection{Data split}

The Leave-One-Subject-Out Cross-Validation (LOSOCV) method was applied to the data of the subjects from the lab. The following impacts on the model were explored by comparing the results: 

\begin{itemize}
	\item Adding another stage for micro labels classification, as an ablation experiment. Correspondingly, another term $\eta L_{CE}^{[2]}$was added to the loss function as in equation~\ref{Loss}, where
	\begin{equation}
		L_{CE}^{[2]}=\frac{1}{TC_1}\sum_{t,c_1}-y_{\text{micro},t,c_1}\log{\hat{y}		_{\text{micro},t,c_1}^{[2]}},
	\label{LCL2}
	\end{equation}
	where $\hat{y}_{\text{micro},t,c_1}^{[2]}$ denotes the predicted micro labels by the additional stage. The results were compared with the proposed model using one stage for micro labels classification.
	\item Different values of loss weight of the first stage ($\eta$).
	\item Different numbers of training examples for micro labels.
	\item Different numbers of stages for macro labels classification.
\end{itemize}

Then, the proposed DS-MS-TCN was compared to the baseline models. Finally, the model was generalized on the home-based data using LOSOCV, with the lab-based data also in the training set.

\subsection{Evaluation}
\label{subsec:evaluation}
\subsubsection{sample-wise evaluation}

For seq-to-seq classification, each input sample yielded a corresponding output. As a result, the sample-wise F1-score was employed to assess the classification performance. The F1-score for class c was calculated as follows:

\begin{equation}
f1_{c}=2*\frac{precision_{c}*recall_{c}}{precision_{c}+recall_{c}},
\label{f:f1}
\end{equation}

where
\begin{equation}
precision_{c}=\frac{TP_{c}}{TP_{c}+FP_{c}}, and
\label{f:precision}
\end{equation}
\begin{equation}
recall_{c}=\frac{TP_{c}}{TP_{c}+FN_{c}}.
\label{f:recell}
\end{equation}

where $TP_{c}$, $FP_{c}$, and $FN_{c}$ denote the numbers of true positive, false positive, and false negative labels from the sample-wise classification output.\par

\subsubsection{segment-wise evaluation}

Given that seq-to-seq classification assigns a label to each sample, the potential for over-segmentation errors exists. This occurs when a continuous activity segment is classified as several smaller segments, with misclassified labels interspersed. As an evaluation metric, segmental F1-scores based on Intersection over Union (IoU) were computed for each class based on Lea et al. \cite{lea_temporal_2017} and Sarapata et al. \cite{10193771}. The Intersection over Union (IoU) matrix was defined as the ratio of the intersection to the union of the predicted and true segments. Initially, a threshold for the IoU values was established. Subsequently, segment-wise true positive ($TPseg$), false positive ($FPseg$), and false negative ($FNseg$) were defined:

\begin{itemize}
	\item $TPseg$: IoU$\geq$threshold
	\item $FPseg$: IoU$<$threshold, true segments shorter than predicted segments
	\item $FNseg$: IoU$<$threshold, true segments longer than predicted segments
\end{itemize}
	
\begin{figure}[!t]
\centering
\includegraphics[width=\columnwidth]{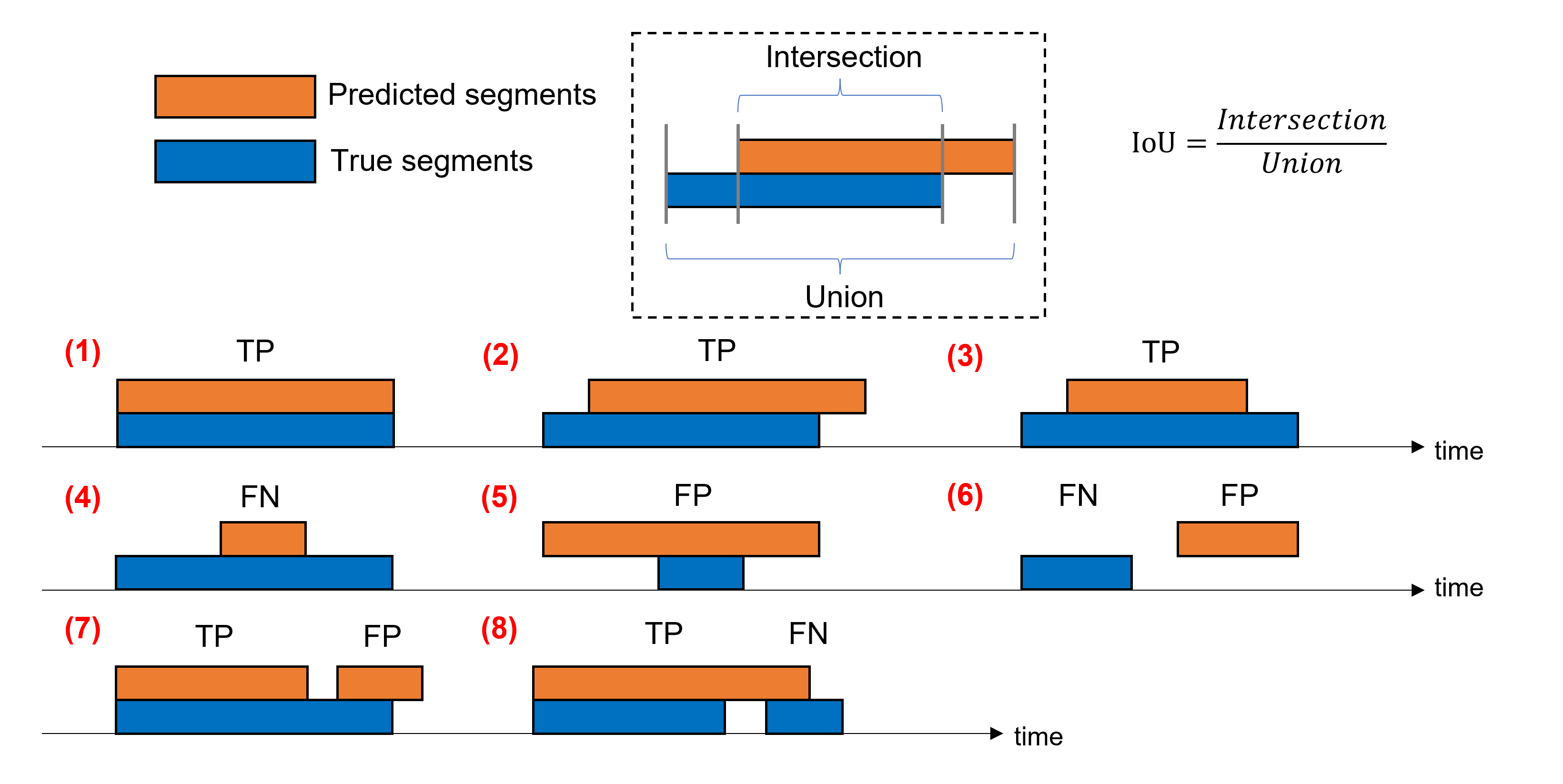}
\caption{The definition of IoU, TP, FP, and FN. There are eight cases shown in the figure. In case 4 and case 5, FN or FP depends on the length of the true and predicted segment. In case 7, if a true segment is predicted as some smaller segments, FP numbers increase. In case 8, if some separate true segments are predicted as one segment, FN numbers increase.}
\label{fig:IoU}
\end{figure}

Fig.~\ref{fig:IoU} illustrates the definition of IoU, $TPseg$, $FPseg$, and $FNseg$ \cite{shang2023otago}. According to equation~\ref{f:f1},~\ref{f:precision}, and~\ref{f:recell}, the segment-wise evaluation matrices could be calculated. Compared with the sample-wise f1-score, this method could also evaluate the over-segmentation errors. In this study, the threshold of 0.5 was applied.

\section{Results}
\label{sec:results}

\subsection{Influence of micro labels stage}

\paragraph{Number of stages for micro labels}
As an ablation experiment, two models were compared: a DS-MS-TCN with two stages for micro labels classification and a DS-MS-TCN with one stage for micro labels classification. Both models applied three stages for macro labels classification. The results are shown in Table~\ref{tab:micro stages}. It is observed that both f1-scores and IoU f1-scores were decreased compared with the model using one stage for micro labels classification. 

\begin{table}[!t]
\caption{F1-scores and IoU f1-scores utilizing one stage and two stages for micro labels classification}
\centering
\label{tab:micro stages}
\begin{tabular}{lllll}
\hline
\multicolumn{5}{c}{f1}                                        \\ \hline
         & ankle plan & abdominal & knee bends & chair rising \\ \hline
1 stage  & 89.777     & 85.363    & 88.298     & 90.028       \\ \hline
2 stages & 88.877     & 77.51     & 82.915     & 84.483       \\ \hline
\multicolumn{5}{c}{IoU f1}                                    \\ \hline
         & ankle plan & abdominal & knee bends & chair rising \\ \hline
1 stage  & 66.667     & 68.852    & 63.333     & 92.308       \\ \hline
2 stages & 64.865     & 60.674    & 57.803     & 73.446       \\ \hline
\end{tabular}
\end{table}

\paragraph{Loss weight of micro labels}

\begin{figure}[!t]
\centering
\subfloat[f1-scores]{\includegraphics[width=\columnwidth]{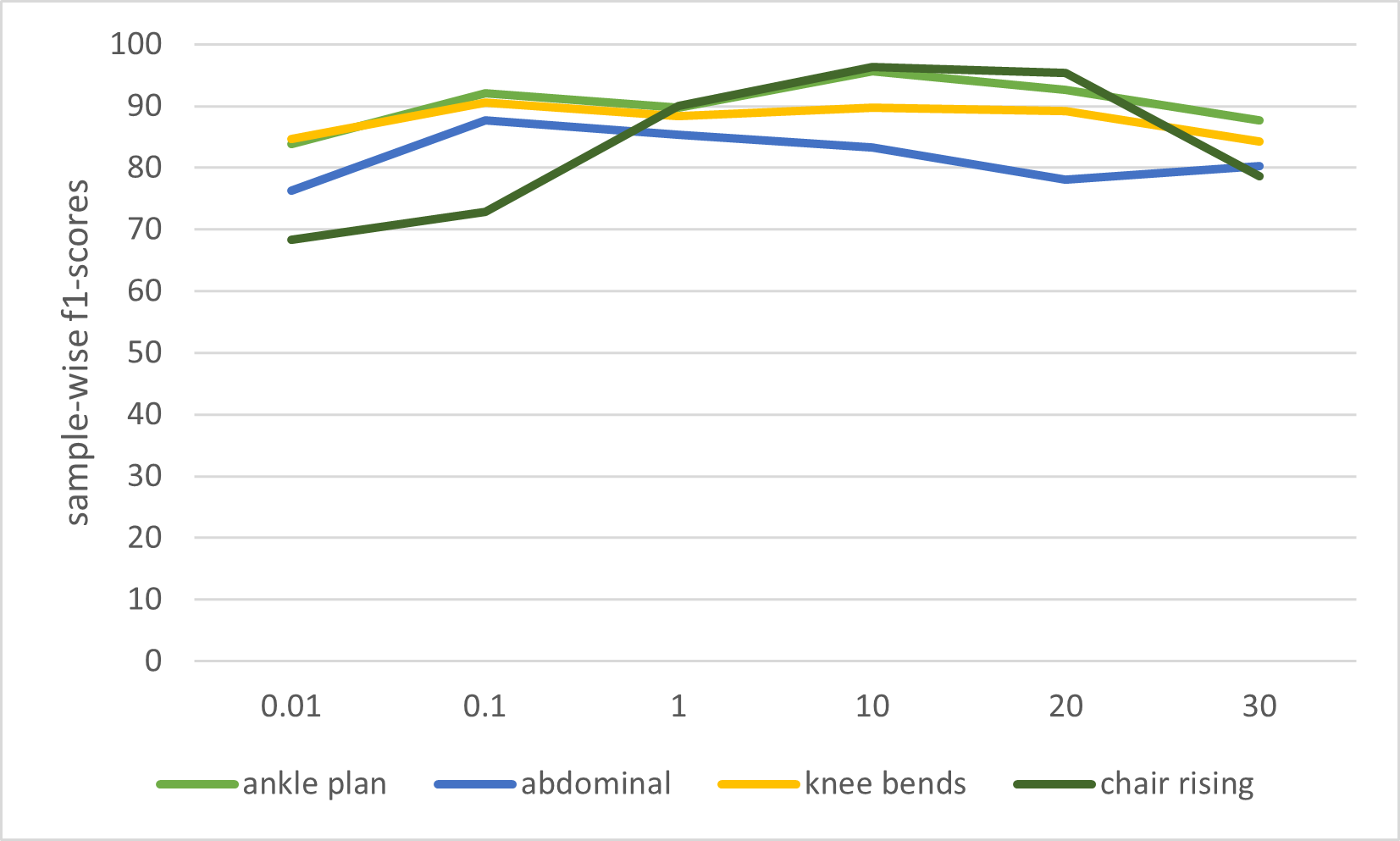}%
}
\hfill
\subfloat[IoU f1-scores]{\includegraphics[width=\columnwidth]{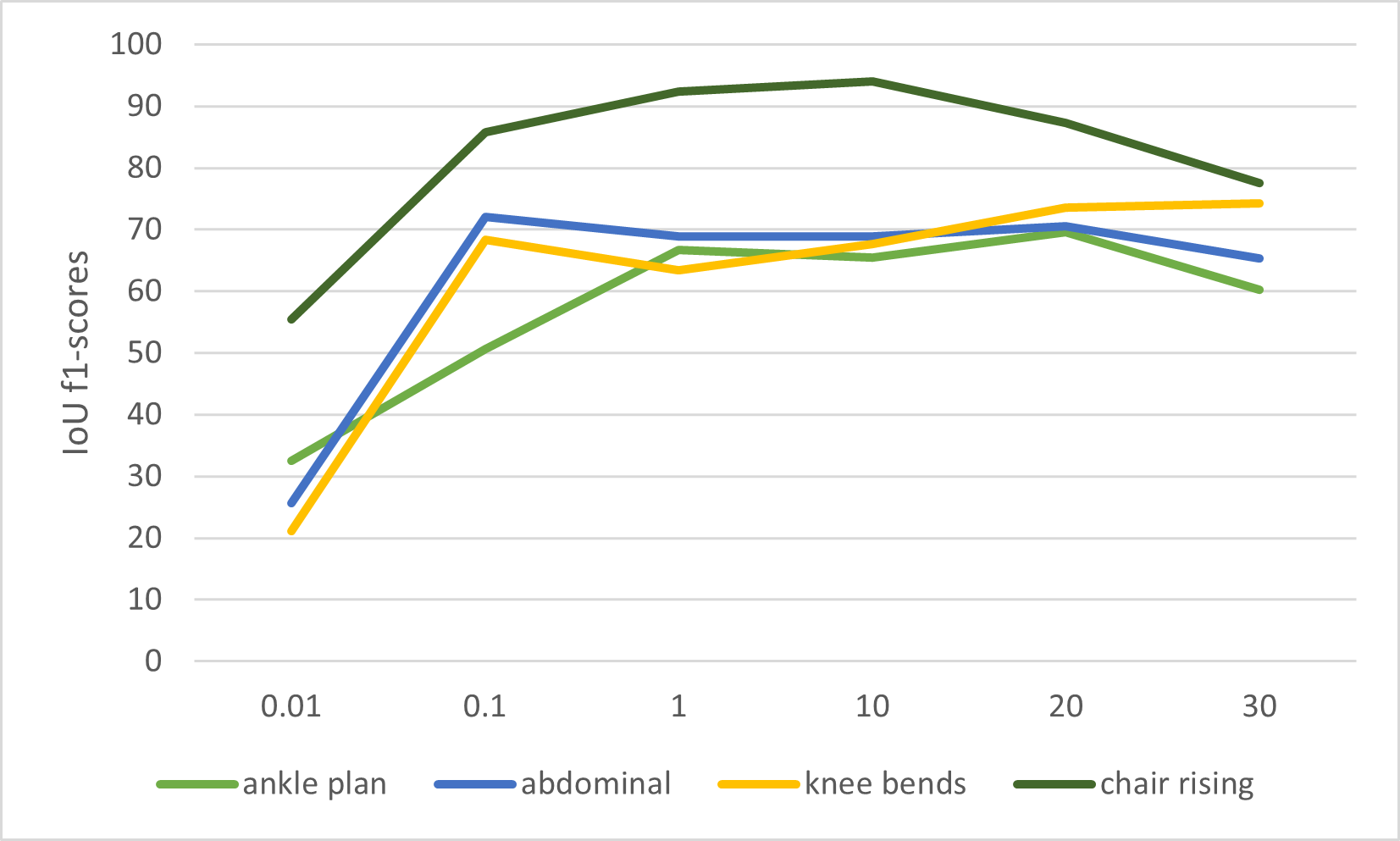}%
}
\caption{F1-scores and IoU f1-scores applying different loss weight for the first stage (i.e. $\eta$)}
\label{fig:loss weight}
\end{figure}

The influence of loss weight of the first stage, i.e. $\eta$ in formula~\ref{Loss}, is shown in Fig.~\ref{fig:loss weight}. The results show that the classification performance for each activity was maximized by different values of $\eta$. For example, the IoU f1-scores reached the maximum at $\eta = 0.1$ for \textit{abdominal muscles} and \textit{knee bends}, whereas for \textit{chair rising} the optimal value for $\eta$ was 10. Besides, the $\eta$ value resulting in the highest f1-scores might not obtain the highest IoU f1-scores. For example, for \textit{chair rising}, the f1-score reached the highest at $\eta = 20$ whereas the IoU f1-score reached the highest at $\eta = 10$. In general, the classification performance was decreased when $\eta$ was lower than 0.1 or higher than 20. Therefore, in the following experiments, the value of $\eta$ was set as default as 1. \par

\paragraph{Number of trained micro labels}

\begin{figure}[!t]
\centering
\includegraphics[width=\columnwidth]{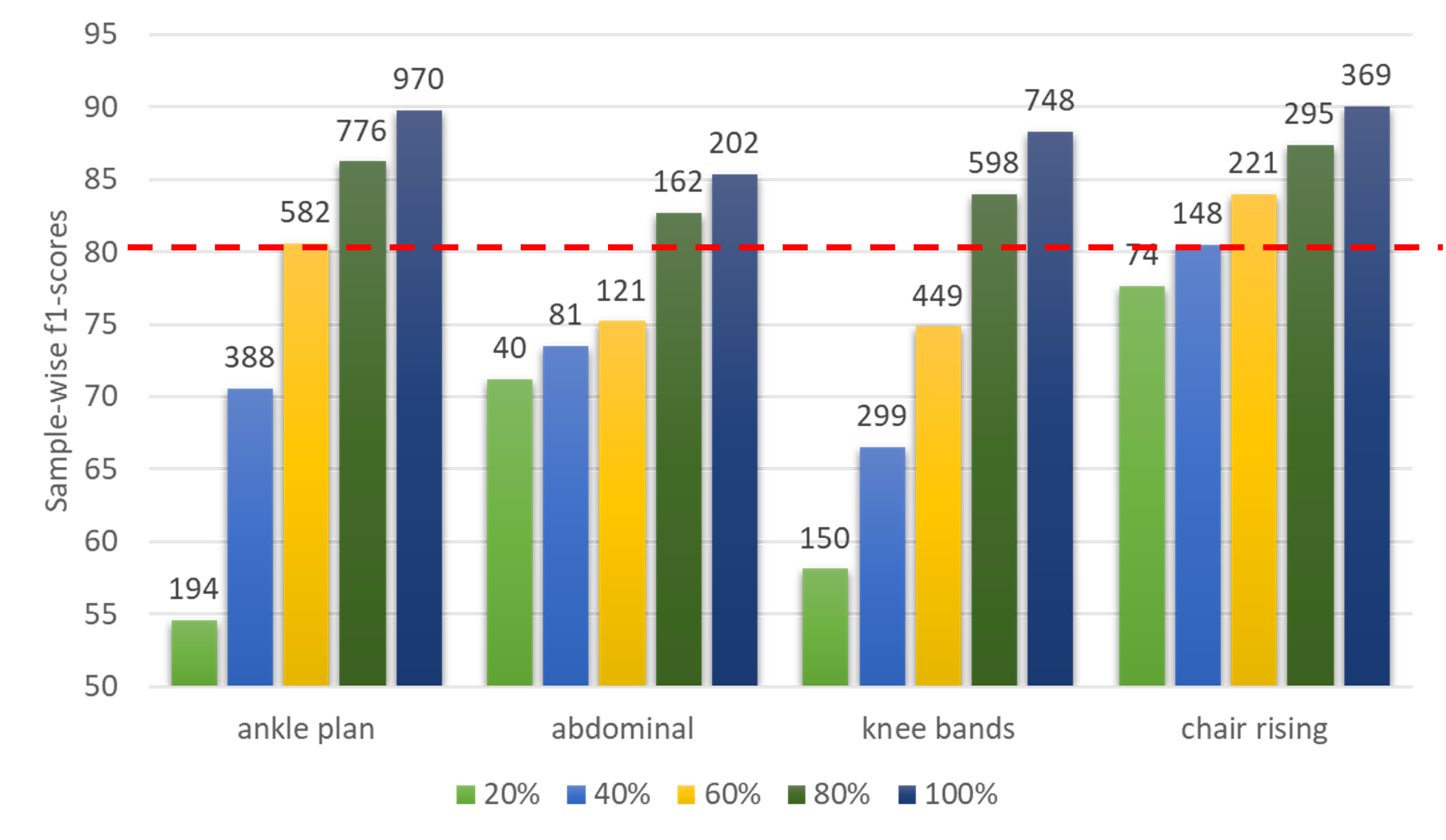}
\caption{F1-scores with different percentages of micro labels in the training set. The red dashed line illustrates a clinically applicable threshold of 80\%. The absolute numbers of labels were shown outside the end of bars.}
\label{fig:micronumber}
\end{figure}

Since manually annotating the micro labels required a huge workload, an experiment was conducted to investigate the impact of the number of micro labels in the training set. Different numbers of micro labels segments were randomly selected to train the model. As shown in Fig.\ref{fig:micronumber}, the f1-scores increase with more percentages of micro labels in the training set. All activities could reach the clinically applicable threshold of 80\% f1-scores when 80\% of micro labels were trained. For \textit{chair rising}, the f1-score could reach the threshold with 40\% micro labels in the training set. However, the bars indicate a continual rise in f1-scores, even when incorporating all micro labels.

\subsection{Influence of macro labels stages}

\begin{figure}[!t]
\centering
\subfloat[f1-scores]{\includegraphics[width=\columnwidth]{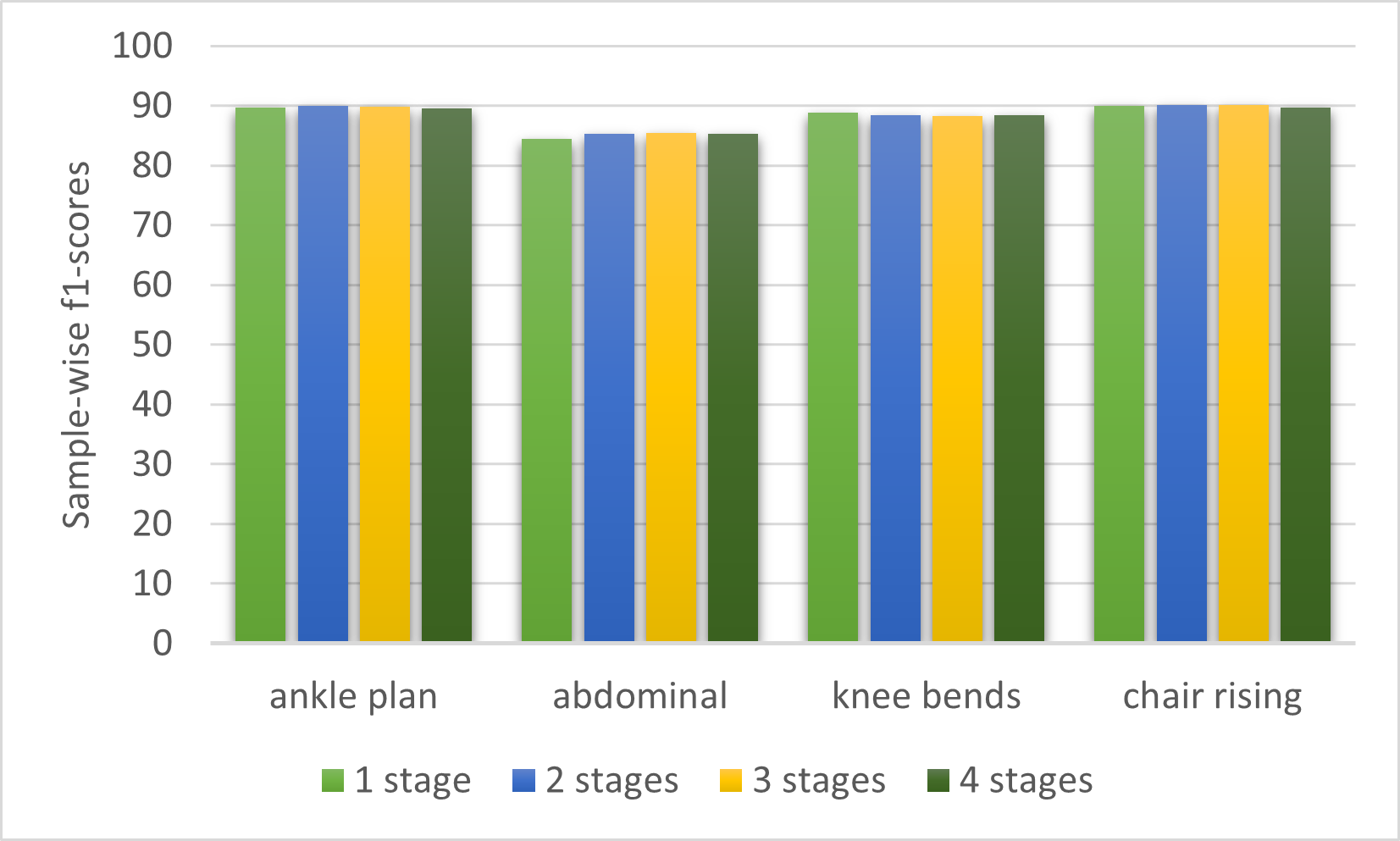}%
\label{fig:stagef1}}
\hfill
\subfloat[IoU f1-scores]{\includegraphics[width=\columnwidth]{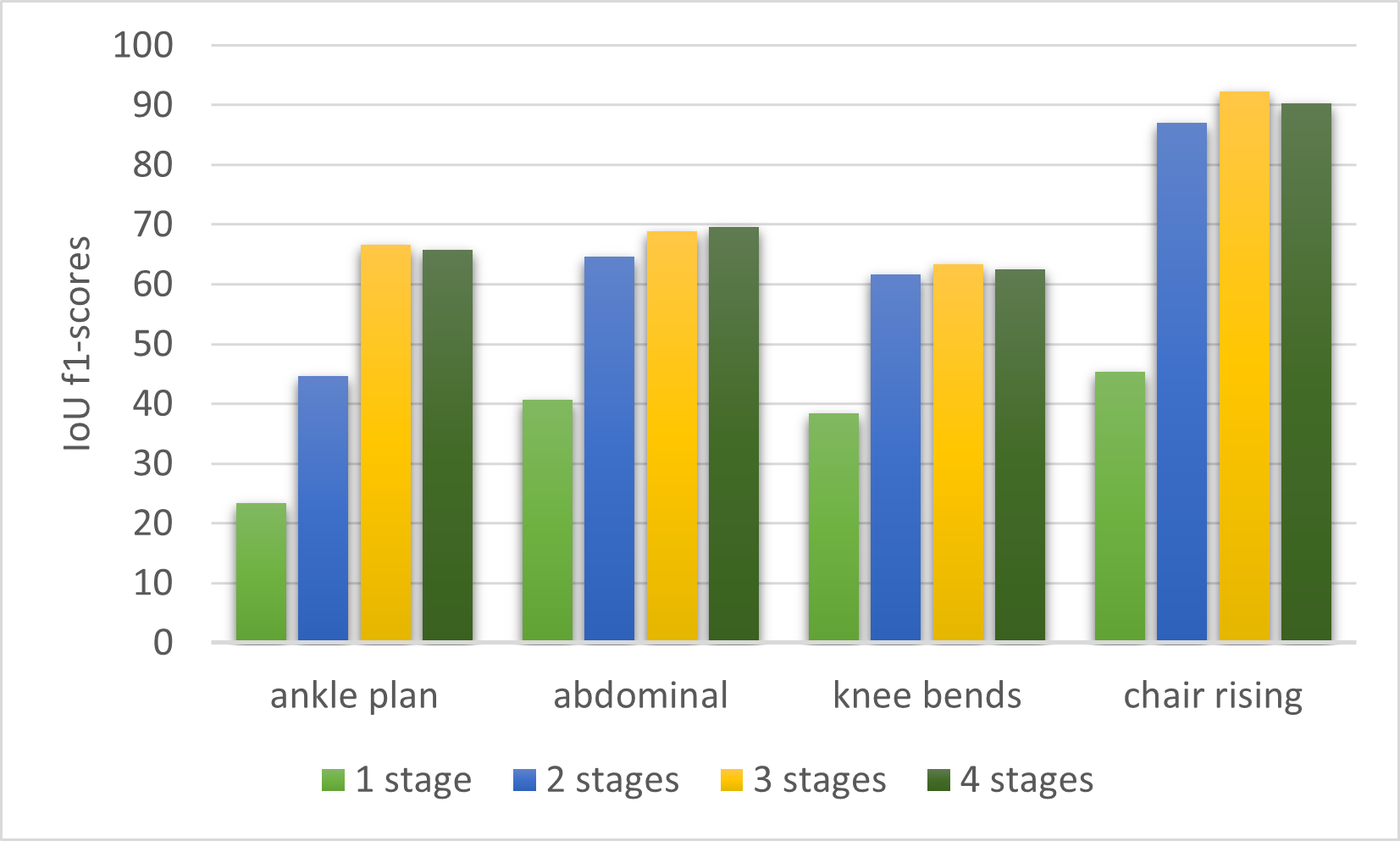}%
\label{fig:stageiouf1}}
\caption{F1-scores and IoU f1-scores applying different number of stages for macro labels classification}
\label{fig:macrostage}
\end{figure}

Fig.~\ref{fig:macrostage} shows the f1-scores and IoU f1-scores of the macro labels applying one, two, three, and four stages in DS-MS-TCN. Fig.~\ref{fig:stagef1} shows that the improvement in f1-scores was not significant by adding the number of stages. On the other hand, the IoU f1-scores were substantially increased for all activities until the third stage. The second stage led to more improvement compared with the third stage, since the third stage did finer tuning of the series, as shown in Fig.~\ref{fig:stageiouf1}. From the fourth stage on, there was no significant improvement anymore. \par

\begin{figure}[!t]
\centering
\includegraphics[width=\columnwidth]{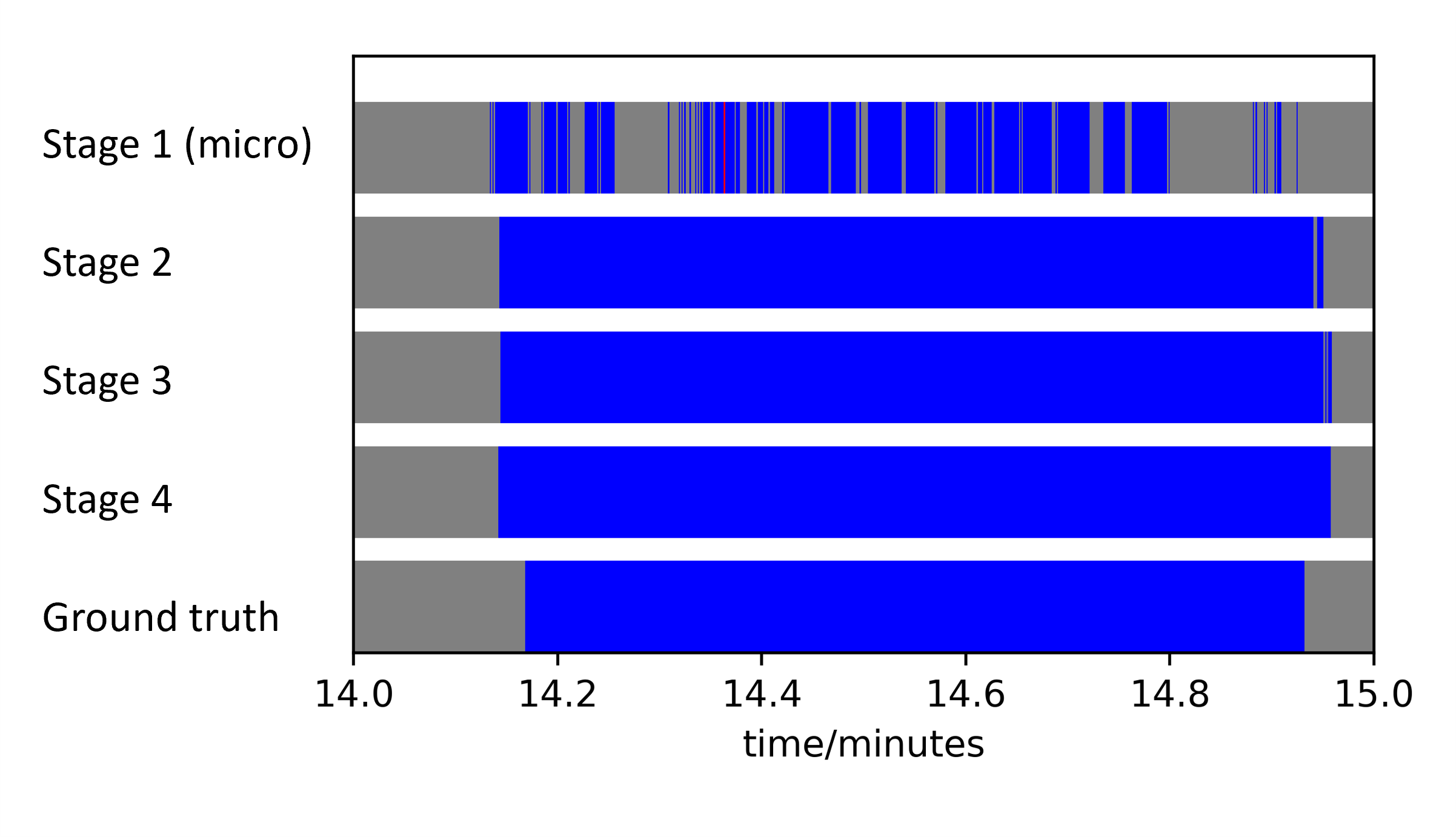}
\caption{The output of each stage from DS-MS-TCN: an example of \textit{ankle plantarflexors}}
\label{fig:visualizestages}
\end{figure}

To visualize the influence of each stage of DS-MS-TCN, Fig.~\ref{fig:visualizestages} shows an example of \textit{ankle plantarflexors} predicted by each stage. The first stage produced micro labels and generated high over-segmentation errors. The second and the third stage produced the macro labels with smaller over-segmentation errors. The final stage then refined the macro labels with the errors removed.

\subsection{Model comparison}
\label{subsec:comparison}

The f1-scores and IoU f1-scores of both the baseline methods and the proposed method are presented in Table~\ref{tab:comparison}.  The hierarchical system, utilizing the sliding window technique, incorporated a post-processing stage to enhance the smoothness of the output sequence \cite{shang2023otago}. Therefore, the IoU f1-scores were higher than the other seq-to-seq baseline models, except the MS-TCN. \par

All baseline models achieved f1-scores exceeding 80\% for \textit{chair rising}. The MS-TCN model (without micro labels) achieved the highest IoU f1-score of 70.27\% for \textit{chair rising}. However, for the other three activities, the IoU f1-scores of the baseline models did not surpass 50\%, despite some reporting f1-scores higher than 80\%. \par

The proposed DS-MS-TCN showed the highest f1-scores over 85\% and IoU f1-scores over 60\% for all activities. It could recognize \textit{chair rising} with both f1-scores and IoU f1-scores over 90\%.

\begin{table*}[!t]
\caption{F1-sores and IoU f1-scores of the proposed DS-MS-TCN and other baseline models}
\centering
\label{tab:comparison}
\begin{tabular}{lllllll}
\hline
\multicolumn{1}{l|}{}                           & \multicolumn{1}{l|}{}                                      &        & ankle plan      & abdominal       & knee bends      & chair rising    \\ \hline
\multicolumn{1}{l|}{\multirow{2}{*}{proposed}}  & \multicolumn{1}{l|}{\multirow{2}{*}{DS-MS-TCN}}            & f1     & \textbf{89.777} & \textbf{85.363} & \textbf{88.298} & \textbf{90.028} \\ \cline{3-7} 
\multicolumn{1}{l|}{}                           & \multicolumn{1}{l|}{}                                      & IoU f1 & \textbf{66.667} & \textbf{68.852} & \textbf{63.333} & \textbf{92.308} \\ \hline
\multicolumn{1}{l|}{\multirow{10}{*}{baseline}} & \multicolumn{1}{l|}{\multirow{2}{*}{MS-TCN}}               & f1     & 72.088          & 66.855          & 44.917          & 83.686          \\ \cline{3-7} 
\multicolumn{1}{l|}{}                           & \multicolumn{1}{l|}{}                                      & IoU f1 & 9.871           & 23.377          & 11.232          & 70.27           \\ \cline{2-7} 
\multicolumn{1}{l|}{}                           & \multicolumn{1}{l|}{\multirow{2}{*}{Transformer}}          & f1     & 74.429          & 72.63           & 59.205          & 84.212          \\ \cline{3-7} 
\multicolumn{1}{l|}{}                           & \multicolumn{1}{l|}{}                                      & IoU f1 & 5.623           & 14.395          & 14.853          & 22.402          \\ \cline{2-7} 
\multicolumn{1}{l|}{}                           & \multicolumn{1}{l|}{\multirow{2}{*}{CNN-LSTM}}             & f1     & 64.42           & 74.745          & 74.918          & 81.817          \\ \cline{3-7} 
\multicolumn{1}{l|}{}                           & \multicolumn{1}{l|}{}                                      & IoU f1 & 3.659           & 16              & 3.902           & 6.015           \\ \cline{2-7} 
\multicolumn{1}{l|}{}                           & \multicolumn{1}{l|}{\multirow{2}{*}{CNN}}                  & f1     & 63.747          & 72.059          & 48.74           & 80.028          \\ \cline{3-7} 
\multicolumn{1}{l|}{}                           & \multicolumn{1}{l|}{}                                      & IoU f1 & 8.791           & 2.492           & 0               & 0.442           \\ \cline{2-7} 
\multicolumn{1}{l|}{}                           & \multicolumn{1}{l|}{\multirow{2}{*}{Hierarchical system*}} & f1     & 78.028          & 85.35           & 77.243          & 84.878          \\ \cline{3-7} 
\multicolumn{1}{l|}{}                           & \multicolumn{1}{l|}{}                                      & IoU f1 & 13.084          & 40.549          & 33.333          & 51.25           \\ \hline
                                                & \multicolumn{6}{l}{*based on the sliding-window technique\cite{shang2023otago}}                                                            
\end{tabular}
\end{table*}

\begin{figure*}[!t]
\centering
\includegraphics[width=1.8\columnwidth]{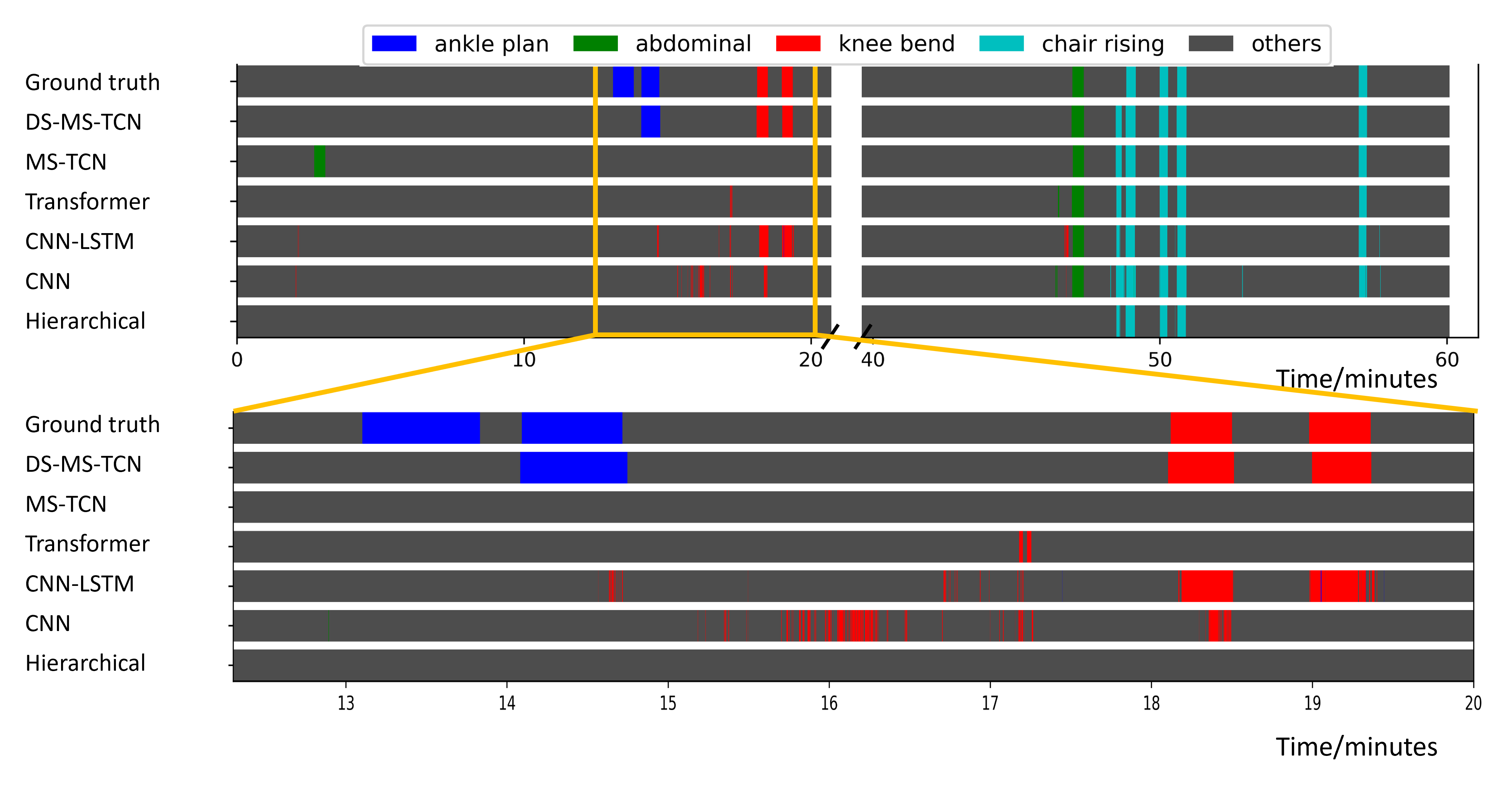}
\caption{The recognized macro labels from a subject. The upper part shows the time series of the whole OEP program. The lower part zooms in the results of \textit{ankle plantarflexors} and \textit{abdominal muscles}}
\label{fig:baseline}
\end{figure*}

Fig.~\ref{fig:baseline} shows the recognized macro labels by different models. The upper part of the figure illustrates the classified time series from a complete OEP program. Some baseline models obtained predicted segments matching the ground truth (e.g. MS-TCN and CNN-LSTM). On the other hand, the other part of the figure at the bottom zooms in the recognition of \textit{ankle plantarflexors} and \textit{abdominal muscles}. The baseline models based on seq-to-seq classification showed many false positive and false negative segments, whereas the proposed DS-MS-TCN model could recognize most segments with only one false positive segment for \textit{chair rising} and one false negative segment for \textit{ankle plantarflexors}.

\subsection{Generalization on subjects at home}

After exploring the optimal model based on the dataset collected in the lab, the home-based dataset was tested using LOSOCV with the lab-based data also in the training set. The f1-scores and IoU f1-scores are shown in Table~\ref{tab:home}. As a comparison, the results of the lab-based dataset by LOSOCV were also in the table. It was observed that both f1-scores and IoU f1-scores decrease when generalizing on the subjects at home. Among the four activities, \textit{ankle plantarflexors} obtained the highest drop, from 89.8\% to 63.4\% for f1-scores, and from 66.7\% to 29.4\% for IoU f1-scores. The IoU f1-scores of \textit{chair rising} also drastically decrease from 92.3\% to 50\%.\par

\begin{table}[!t]
\caption{F1-scores and IoU f1-scores of the lab-based data and home-based data}
\centering
\label{tab:home}
\begin{tabular}{lllll}
\hline
\multicolumn{5}{c}{Lab}                                     \\ \hline
       & ankle plan & abdominal & keen bends & chair rising \\ \hline
f1     & 89.777     & 85.363    & 88.298     & 90.028       \\ \hline
IoU f1 & 66.667     & 68.852    & 63.333     & 92.308       \\ \hline
\multicolumn{5}{c}{Home}                                    \\ \hline
       & ankle plan & abdominal & keen bends & chair rising \\ \hline
f1     & 63.401     & 87.377    & 67.570     & 85.909       \\ \hline
IoU f1 & 29.412     & 44.444    & 16.667     & 50.000       \\ \hline
\end{tabular}
\end{table}

\begin{figure}[!t]
\centering
\includegraphics[width=\columnwidth]{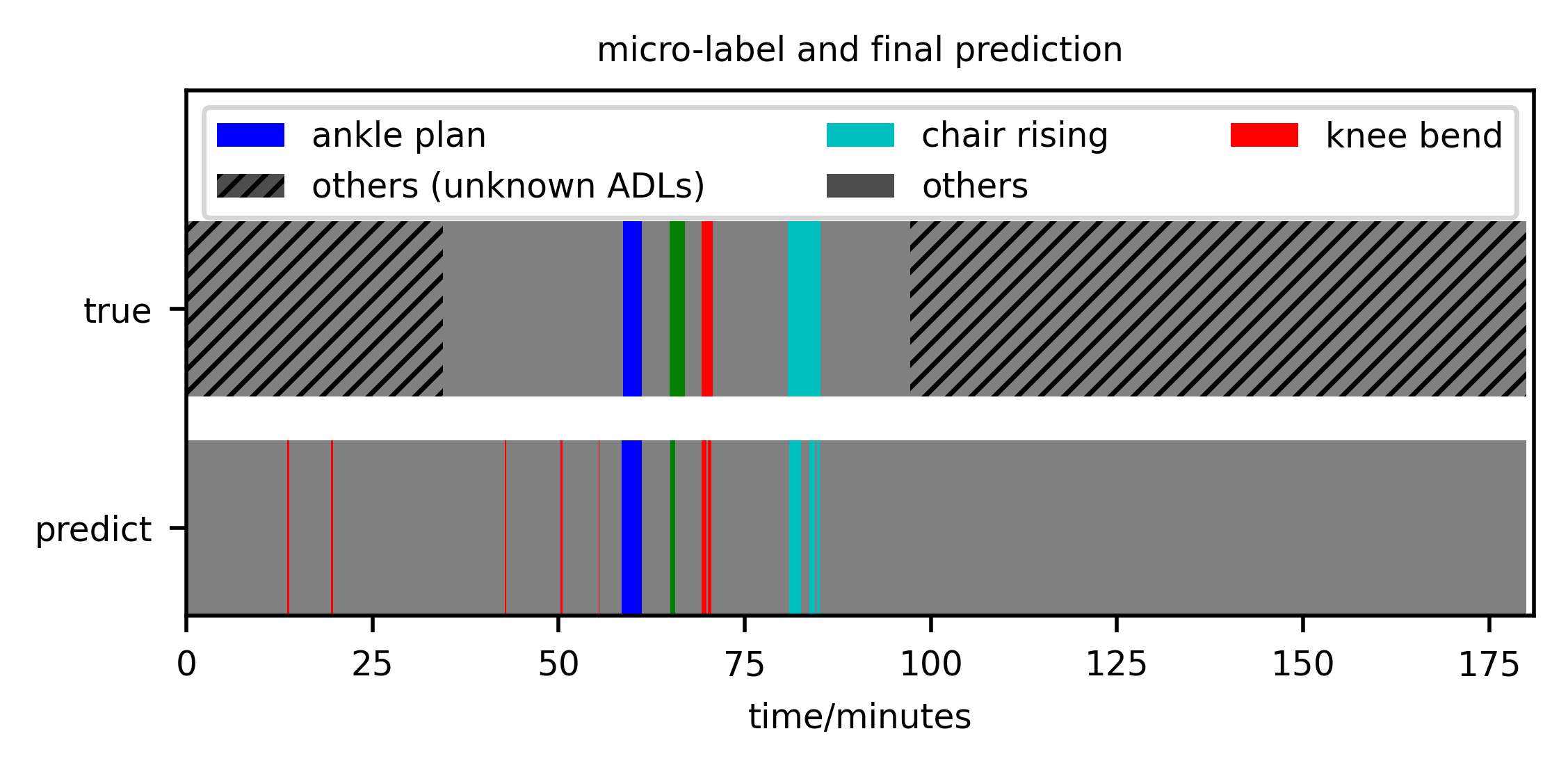}
\caption{The recognized macro labels from a subject at home. The shadowed area illustrates the ADLs that were not observed in daily life due to the limitation of camera installation.}
\label{fig:home}
\end{figure}

An example of predicted labels for one subject at home is shown in Fig.~\ref{fig:home}. The shadowed blocks highlight the labels of ADLs that were not observed by the camera. The results show that the model generated several false positive segments for \textit{knee bends}, whereas the other labels of ADLs were correctly predicted. However, there were more over-segmentation errors during the OEP. On the one hand, a single segment of macro activity was classified as more smaller segments. On the other hand, some other activities were classified as short false positive macro labels.

\subsection{Comparison with the previous studies}

This study is a follow-up study of the previous work \cite{shang2023otago}, which has been discussed in Section~\ref{subsec:comparison}. Besides, there were only two studies applying wearable IMUs for Otago exercises recognition. One study \cite{dedeyne_exploring_2021} predicted ADLs and OEP (binary classification) and obtained the f1-score of 77.7\%. The other study \cite{bevilacqua_human_2019} explored most Otago sub-exercises using different sensor combinations, as shown in Table~\ref{tab:previous}. To compare with this study, the table only presents the results including the waist-mounted IMU. The previous study was performed in a highly controlled environment (young healthy adults in the lab). It applied CNN based on the sliding window technique (seq-to-one) evaluated based on 5-fold cross-validation. This study reported higher f1-scores for \textit{knee bends} and \textit{chair rising} even without the IMUs on the shank and feet. However, the previous study reported higher f1-scores for \textit{ankle plantarflexors} with two more IMUs.\par

\begin{table*}[!t]
\caption{Comparison with a previous study. The f1-scores of the previous study was based on 5-fold cross-validation, whereas this study used LOSOCV. \textit{Abdominal muscles} was not collected in the previous study.}
\label{tab:previous}
\centering
\begin{tabular}{llllllll}
\hline
\multirow{2}{*}{Study}       & \multirow{2}{*}{sensors}                                                            & \multirow{2}{*}{subjects}                                                           & \multirow{2}{*}{scenario} & \multirow{2}{*}{method}                                      & \multicolumn{3}{c}{f1-scores}            \\ \cline{6-8} 
                             &                                                                                     &                                                                                     &                           &                                                              & ankle plan  & knee bends  & chair rising \\ \hline
\cite{bevilacqua_human_2019} & 1 IMU (waist)                                                                       & 19 healthy adults                                                                   & lab                       & \begin{tabular}[c]{@{}l@{}}CNN,\\ seq-to-one\end{tabular}    & 32          & 31          & 49           \\ \hline
\cite{bevilacqua_human_2019} & \begin{tabular}[c]{@{}l@{}}3 IMU \\ (right shank + right feet + waist)\end{tabular} & 19 healthy adults                                                                   & lab                       & \begin{tabular}[c]{@{}l@{}}CNN,\\ seq-to-one\end{tabular}    & \textbf{98} & 84          & 67           \\ \hline
\cite{bevilacqua_human_2019} & \begin{tabular}[c]{@{}l@{}}3 IMU \\ (left shank + left feet + waist)\end{tabular}   & 19 healthy adults                                                                   & lab                       & \begin{tabular}[c]{@{}l@{}}CNN,\\ seq-to-one\end{tabular}    & \textbf{98} & 85          & 71           \\ \hline
This study                   & 1 IMU (waist)                                                                       & \begin{tabular}[c]{@{}l@{}}35 older adults, \\ some of them sarcopenic\end{tabular} & lab                       & \begin{tabular}[c]{@{}l@{}}DS-MS-TCN,\\ seq-to-seq\end{tabular} & 90          & \textbf{88} & \textbf{90}  \\ \hline
This study                   & 1 IMU (waist)                                                                       & \begin{tabular}[c]{@{}l@{}}7 older adults, \\ some of them sarcopenic\end{tabular}  & home                      & \begin{tabular}[c]{@{}l@{}}DS-MS-TCN\\ seq-to-seq\end{tabular}  & 49          & 83          & 80           \\ \hline
\end{tabular}
\end{table*}

\section{Discussion}
\label{sec:discussion}

\subsection{Seq-to-seq HAR}

Seq-to-seq models for activity recognition have not been explored much compared with the sliding window technique. In the previous study \cite{shang2023otago}, a large sliding window was first applied to distinguish the ADLs and OEP (binary classification). Then, a small sliding window was applied to classify which exercises from OEP were performed. The drawback of this technique was that the sizes of sliding windows had to be determined and the post-processing stages were necessary. Also, post-processing was necessary to improve the smoothness of the output, because the predicted segments were independent from each other. It is also notable that this study only explored four types of OEP activities. This decision was influenced by the fact that some exercises lacked clinical significance, and some were challenging to identify due to limitations in sensor capabilities.  \par

With seq-to-seq techniques, the models could learn temporal dependencies from a longer time series. Therefore, it could recognize the specific exercises using one model, rather than two hierarchical models. MS-TCN and DS-MS-TCN applied multiple convolutional layers with different dilation factors. Therefore, both short-term and long-term information could be captured. Also, with the TMSE functions, the output was smoothed by considering the adjacent samples. The results in Fig.\ref{fig:home} show limited confusion between ADLs and Otago exercises, despite most training samples belonging to \textit{others}, as shown in Table~\ref{tab:duration}.\par

\subsection{DS-MS-TCN based on micro labels}

This study proposes an innovative annotation method for activity recognition using DS-MS-TCN. In the first stage of the model, each repetition of the activities (micro activity) was recognized. The reason for using micro labels was to bring less irrelevant information to the model. Hence, micro labels contained lower variance. For example, when performing \textit{ankle plantarflexors}, subjects paused between each two repetitions. During these pauses, the sensors did not capture any information whereas they were still labeled as positive. Although the pauses were short (no more than one second), for seq-to-seq models, it was harder to learn and generalize. Therefore, DS-MS-TCN based on micro labels outperformed MS-TCN with the same structure. \par

Due to the lower variance of micro labels, smaller training sets were needed to get good performance. Compared with micro labels that were abundant, macro labels are less prominent in the data. Using the duel-scale model, from micro to macro has to do with distributions of micro labels rather than the underlying sensor signals. Therefore, the performance of macro activity classification was improved using the same dataset.\par

The recognition performance of micro labels was not expected to be perfect because each micro labels segment was too short. However, introducing an additional stage for micro labels classification did not yield improvements in the results. This can be attributed to the refinement process, which resulted in the elimination of certain short micro labels and contributed to the loss of information for macro label classification. \par

The loss weight of the first stage ($\eta$) had varying effects on the final f1-scores for different activities. Therefore, it was difficult to determine $\eta$ that led to the highest overall f1-scores and IoU f1-scores. However, the influence of $\eta$  was rather constant within a certain range (between 0.1 and 20). It was thus set as one in this study. \par

Some micro activities happen in both Otago exercises and other ADLs. However, the activities happening in ADLs should not be classified as Otago exercises. The difference among both was that Otago exercises were always performed consistently with multiple repetitions whereas the ADLs were performed for a single repetition. Therefore, the first stage could not recognize the difference and the second stage was necessary to remove over-segmentation error. If there was only a single repetition of micro activities, it was excluded from the Otago exercises by the higher stages. For similar reasons, the third stage and fourth stage were also applied, since the second stage was not sufficient to completely remove the over-segmentation error.\par

Although Fig.~\ref{fig:micronumber} shows that 80\% of the micro labels were enough to reach the f1-scores of 80\%, the f1-scores were not maximized even with all micro labels used. With the more variance of the activities, the number of micro labels had more influence on the results. For example, \textit{chair rising} and \textit{abdominal muscles} had more distinguishable and consistent IMU signals and hence less variance. Thus they required fewer micro labels to reach the threshold. On the other hand, \textit{ankle plantarflexors} and \textit {knee bends} required more micro labels to cover the higher variance of the IMU signals. \par

The annotation of micro labels required a substantial workload. Consequently, it becomes crucial to determine the minimum number of micro labels that result in optimal classification performance. In Fig. \ref{fig:micronumber}, theoretically, the F1-scores should plateau after involving a certain quantity of micro labels. The limitation of this study lies in the scarcity of micro labels, preventing exploration of the optimal number for achieving peak performance.

\subsection{Otago exercises recognition}

This study proposes a system with the simplest and most user-friendly implementation using only one wearable IMU on the waist. Therefore, only the activities involving trunk movement could be recognized. The four activities explored in this study were all important for strength (\textit{ankle plantarflexors} and \textit{abdominal muscles})and balance (\textit{knee bends} and \textit{chair rising}). Compared with the previous study \cite{bevilacqua_human_2019}, the proposed system was validated in a less controlled environment focusing on older adults. When comparing the results using one IMU on the waist, the proposed system reported higher f1-scores, proving the advantages of the seq-to-seq DS-MS-TCN model. This study also proved the possibility of recognizing these activities without additional IMU. \par

The previous study \cite{bevilacqua_human_2019} reported lower f1-scores for \textit{chair rising} because it was confused with \textit{knee bends}. For younger healthy adults, these two activities are similar regarding the changing of vertical acceleration. However, in this study, these two activities were more distinguishable for older adults because they did not have support for \textit{knee bends} whereas they had a chair to sit on for \textit{chair rising}. The vertical acceleration was thus different  for these two activities. \par

Generalization on the subjects at home showed lower f1-scores than the subjects in the lab. One of the reasons might be that subjects at home only followed a booklet and did not receive instructions from the researchers. Therefore, the exercises that they performed were not as standard as the subjects in the lab. Another reason was that the other activities were more complicated and could lead to confusion. In the future study, more data will be collected in daily life to capture more variance of the signals. And the model should be improved regarding generalization ability.\par

\section{Conclusion}
\label{sec:conclusion}

This paper proposes a system designed for older adults to recognize OEP in their daily lives. The DS-MS-TCN model demonstrates the efficacy of micro labels classification in enhancing overall classification performance. This dual-scale system contributes novel perspectives to the field of human activity recognition.\par

Future research directions include improving the recognition of micro labels to enable quantitative analysis of exercise repetitions within each OEP routine. Considering the substantial workload for micro labels annotation, transfer learning or semi-supervised learning could be applied with the limited number of labels \cite{oh2021study}. Additionally, the integration of more tools, such as video-based classification systems, could contribute to enhancing the overall classification performance.

\section*{Acknowledgment}

This work was supported by the China Scholarship Council (CSC).\par

The ENHANce project (S60763) received a junior research project grant from the Research Foundation Flanders (FWO) (G099721N). The funding provider did not contribute or influence the design of the study and data collection, analysis and interpretation in writing this manuscript.

\section*{REFERENCES}

\bibliographystyle{ieeetr}
\bibliography{generic-color}

\begin{thebibliography}{10}

\bibitem{wang_survey_2019}
Y.~Wang, S.~Cang, and H.~Yu, ``A survey on wearable sensor modality centred
  human activity recognition in health care,'' {\em Expert Systems with
  Applications}, vol.~137, pp.~167--190, Dec. 2019.

\bibitem{mekruksavanich2022multimodal}
S.~Mekruksavanich and A.~Jitpattanakul, ``Multimodal wearable sensing for
  sport-related activity recognition using deep learning networks,'' {\em
  Journal of Advances in Information Technology}, 2022.

\bibitem{bianchi2019iot}
V.~Bianchi, M.~Bassoli, G.~Lombardo, P.~Fornacciari, M.~Mordonini, and
  I.~De~Munari, ``Iot wearable sensor and deep learning: An integrated approach
  for personalized human activity recognition in a smart home environment,''
  {\em IEEE Internet of Things Journal}, vol.~6, no.~5, pp.~8553--8562, 2019.

\bibitem{thomas_does_2010}
S.~Thomas, S.~Mackintosh, and J.~Halbert, ``Does the ‘{Otago} exercise
  programme’ reduce mortality and falls in older adults?: a systematic review
  and meta-analysis,'' {\em Age and Ageing}, vol.~39, pp.~681--687, Nov. 2010.

\bibitem{mat_effect_2018}
S.~Mat, C.~T. Ng, P.~J. Tan, N.~Ramli, F.~Fadzli, F.~I. Rozalli, M.~Mazlan,
  K.~D. Hill, and M.~P. Tan, ``Effect of {Modified} {Otago} {Exercises} on
  {Postural} {Balance}, {Fear} of {Falling}, and {Fall} {Risk} in {Older}
  {Fallers} {With} {Knee} {Osteoarthritis} and {Impaired} {Gait} and {Balance}:
  {A} {Secondary} {Analysis},'' {\em PM\&R}, vol.~10, no.~3, pp.~254--262,
  2018.

\bibitem{almarzouki_improved_2020}
R.~Almarzouki, G.~Bains, E.~Lohman, B.~Bradley, T.~Nelson, S.~Alqabbani,
  A.~Alonazi, and N.~Daher, ``Improved balance in middle-aged adults after 8
  weeks of a modified version of otago exercise program: A randomized
  controlled trial,'' {\em Plos one}, vol.~15, no.~7, p.~e0235734, 2020.

\bibitem{shang2023otago}
M.~Shang, L.~Dedeyne, J.~Dupont, L.~Vercauteren, N.~Amini, L.~Lapauw,
  E.~Gielen, S.~Verschueren, C.~Varon, W.~De~Raedt, and B.~Vanrumste, ``Otago
  exercises monitoring for older adults by a single imu and hierarchical
  machine learning models,'' {\em IEEE Transactions on Neural Systems and
  Rehabilitation Engineering}, vol.~32, pp.~462--471, 2024.

\bibitem{dang2020sensor}
L.~M. Dang, K.~Min, H.~Wang, M.~J. Piran, C.~H. Lee, and H.~Moon,
  ``Sensor-based and vision-based human activity recognition: A comprehensive
  survey,'' {\em Pattern Recognition}, vol.~108, p.~107561, 2020.

\bibitem{kyritsis2017food}
K.~Kyritsis, C.~Diou, and A.~Delopoulos, ``Food intake detection from inertial
  sensors using lstm networks,'' in {\em New Trends in Image Analysis and
  Processing--ICIAP 2017: ICIAP International Workshops, WBICV, SSPandBE, 3AS,
  RGBD, NIVAR, IWBAAS, and MADiMa 2017, Catania, Italy, September 11-15, 2017,
  Revised Selected Papers 19}, pp.~411--418, Springer, 2017.

\bibitem{9153742}
H.~Bi, M.~Perello-Nieto, R.~Santos-Rodriguez, and P.~Flach, ``Human activity
  recognition based on dynamic active learning,'' {\em IEEE Journal of
  Biomedical and Health Informatics}, vol.~25, no.~4, pp.~922--934, 2021.

\bibitem{li_human_2022}
Y.~Li and L.~Wang, ``Human activity recognition based on residual network and
  bilstm,'' {\em Sensors}, vol.~22, no.~2, p.~635, 2022.

\bibitem{ellis_multi-sensor_2014}
K.~Ellis, J.~Kerr, S.~Godbole, and G.~Lanckriet, ``Multi-sensor physical
  activity recognition in free-living,'' in {\em Proceedings of the 2014 {ACM}
  {International} {Joint} {Conference} on {Pervasive} and {Ubiquitous}
  {Computing}: {Adjunct} {Publication}}, (Seattle Washington), pp.~431--440,
  ACM, Sept. 2014.

\bibitem{farha_ms-tcn_2019}
Y.~A. Farha and J.~Gall, ``Ms-tcn: Multi-stage temporal convolutional network
  for action segmentation,'' in {\em Proceedings of the IEEE/CVF Conference on
  Computer Vision and Pattern Recognition (CVPR)}, June 2019.

\bibitem{wang2022eat}
C.~Wang, T.~S. Kumar, W.~De~Raedt, G.~Camps, H.~Hallez, and B.~Vanrumste,
  ``Eat-radar: Continuous fine-grained eating gesture detection using fmcw
  radar and 3d temporal convolutional network,'' {\em arXiv preprint
  arXiv:2211.04253}, 2022.

\bibitem{shang2023multi}
M.~Shang, C.~De~Bleecker, J.~Vanrenterghem, R.~De~Ridder, S.~Verschueren,
  C.~Varon, W.~De~Raedt, and B.~Vanrumste, ``A multi-stage temporal
  convolutional network for volleyball jumps classification using a
  waist-mounted imu,'' {\em arXiv preprint arXiv:2310.13097}, 2023.

\bibitem{zhao_deep_2018}
Y.~Zhao, R.~Yang, G.~Chevalier, X.~Xu, and Z.~Zhang, ``Deep {Residual}
  {Bidir}-{LSTM} for {Human} {Activity} {Recognition} {Using} {Wearable}
  {Sensors},'' {\em Mathematical Problems in Engineering}, vol.~2018,
  p.~e7316954, Dec. 2018.
\newblock Publisher: Hindawi.

\bibitem{8684824}
J.~Huang, S.~Lin, N.~Wang, G.~Dai, Y.~Xie, and J.~Zhou, ``Tse-cnn: A two-stage
  end-to-end cnn for human activity recognition,'' {\em IEEE Journal of
  Biomedical and Health Informatics}, vol.~24, no.~1, pp.~292--299, 2020.

\bibitem{lee_human_2017}
S.-M. Lee, S.~M. Yoon, and H.~Cho, ``Human activity recognition from
  accelerometer data using {Convolutional} {Neural} {Network},'' in {\em 2017
  {IEEE} {International} {Conference} on {Big} {Data} and {Smart} {Computing}
  ({BigComp})}, pp.~131--134, Feb. 2017.
\newblock ISSN: 2375-9356.

\bibitem{wagner_activity_2017}
D.~Wagner, K.~Kalischewski, J.~Velten, and A.~Kummert, ``Activity recognition
  using inertial sensors and a 2-{D} convolutional neural network,'' in {\em
  2017 10th {International} {Workshop} on {Multidimensional} ({nD}) {Systems}
  ({nDS})}, pp.~1--6, Sept. 2017.

\bibitem{tao2018worker}
W.~Tao, Z.-H. Lai, M.~C. Leu, and Z.~Yin, ``Worker activity recognition in
  smart manufacturing using imu and semg signals with convolutional neural
  networks,'' {\em Procedia Manufacturing}, vol.~26, pp.~1159--1166, 2018.

\bibitem{mutegeki_cnn-lstm_2020}
R.~Mutegeki and D.~S. Han, ``A {CNN}-{LSTM} {Approach} to {Human} {Activity}
  {Recognition},'' in {\em 2020 {International} {Conference} on {Artificial}
  {Intelligence} in {Information} and {Communication} ({ICAIIC})},
  pp.~362--366, Feb. 2020.

\bibitem{mekruksavanich_smartwatch-based_2020}
S.~Mekruksavanich and A.~Jitpattanakul, ``Smartwatch-based {Human} {Activity}
  {Recognition} {Using} {Hybrid} {LSTM} {Network},'' in {\em 2020 {IEEE}
  {SENSORS}}, pp.~1--4, Oct. 2020.
\newblock ISSN: 2168-9229.

\bibitem{dirgova2022wearable}
I.~Dirgov{\'a}~Lupt{\'a}kov{\'a}, M.~Kubov{\v{c}}{\'\i}k, and
  J.~Posp{\'\i}chal, ``Wearable sensor-based human activity recognition with
  transformer model,'' {\em Sensors}, vol.~22, no.~5, p.~1911, 2022.

\bibitem{trujillo2023accuracy}
M.~F. Trujillo-Guerrero, S.~Rom{\'a}n-Niemes, M.~Ja{\'e}n-Vargas, A.~Cadiz,
  R.~Fonseca, and J.~J. Serrano-Olmedo, ``Accuracy comparison of cnn, lstm, and
  transformer for activity recognition using imu and visual markers,'' {\em
  IEEE Access}, 2023.

\bibitem{dedeyne_exploring_2021}
L.~Dedeyne, J.~A. Wullems, J.~Dupont, J.~Tournoy, E.~Gielen, and
  S.~Verschueren, ``Exploring machine learning models based on accelerometer
  sensor alone or combined with gyroscope to classify home-based exercises and
  physical behavior in (pre) sarcopenic older adults,'' {\em Journal for the
  Measurement of Physical Behaviour}, vol.~4, no.~2, pp.~174--186, 2021.

\bibitem{cruz2010sarcopenia}
A.~J. Cruz-Jentoft, J.~P. Baeyens, J.~M. Bauer, Y.~Boirie, T.~Cederholm,
  F.~Landi, F.~C. Martin, J.-P. Michel, Y.~Rolland, S.~M. Schneider, {\em
  et~al.}, ``Sarcopenia: European consensus on definition and diagnosisreport
  of the european working group on sarcopenia in older peoplea. j. cruz-gentoft
  et al.,'' {\em Age and ageing}, vol.~39, no.~4, pp.~412--423, 2010.

\bibitem{lea_temporal_2017}
C.~Lea, M.~D. Flynn, R.~Vidal, A.~Reiter, and G.~D. Hager, ``Temporal
  {Convolutional} {Networks} for {Action} {Segmentation} and {Detection},'' in
  {\em 2017 {IEEE} {Conference} on {Computer} {Vision} and {Pattern}
  {Recognition} ({CVPR})}, (Honolulu, HI), pp.~1003--1012, IEEE, July 2017.

\bibitem{10193771}
G.~Sarapata, Y.~Dushin, G.~Morinan, J.~Ong, S.~Budhdeo, B.~Kainz, and
  J.~O'Keeffe, ``Video-based activity recognition for automated motor
  assessment of parkinson's disease,'' {\em IEEE Journal of Biomedical and
  Health Informatics}, vol.~27, no.~10, pp.~5032--5041, 2023.

\bibitem{bevilacqua_human_2019}
A.~Bevilacqua, K.~MacDonald, A.~Rangarej, V.~Widjaya, B.~Caulfield, and
  T.~Kechadi, ``Human activity recognition with convolutional neural
  networks,'' in {\em Joint European Conference on Machine Learning and
  Knowledge Discovery in Databases}, pp.~541--552, Springer, 2019.

\bibitem{oh2021study}
S.~Oh, A.~Ashiquzzaman, D.~Lee, Y.~Kim, and J.~Kim, ``Study on human activity
  recognition using semi-supervised active transfer learning,'' {\em Sensors},
  vol.~21, no.~8, p.~2760, 2021.

\end{thebibliography}

\end{document}